%% file: main.tex
\documentclass[10pt,twocolumn,letterpaper]{article}

\usepackage{iccv}
\usepackage{times}
\usepackage{epsfig}
\usepackage{graphicx}
\usepackage{amsmath}
\usepackage{amssymb}
\usepackage{mathtools}
\usepackage{multirow}
\usepackage{enumitem}
\usepackage{booktabs}

\usepackage[pagebackref=true,breaklinks=true,letterpaper=true,colorlinks,bookmarks=false]{hyperref}

\makeatletter
\def\blfootnote{\xdef\@thefnmark{}\@footnotetext}
\makeatother

\iccvfinalcopy 

\def\iccvPaperID{6} 

\ificcvfinal\fi

\graphicspath{{figures/}}

\input{sec/notation}

\DeclareGraphicsExtensions{.jpeg,.jpg,.png}

\begin{document}

\title{Learned Semantic Multi-Sensor Depth Map Fusion}

\author{Denys Rozumnyi$^{1,3}$
\and
Ian Cherabier$^{1}$
\and
Marc Pollefeys$^{1,2}$
\and
Martin R. Oswald$^{1}$ 
\and
{\normalsize $^{1}$Department of Computer Science, ETH Zurich }
\and
{\normalsize  $^{2}$Microsoft }
\and
{\normalsize $^{3}$Visual Recognition Group, Faculty of Electrical Engineering, Czech Technical University in Prague, Czech Republic }
}

\maketitle

\begin{abstract}
\input{sec/abstract}
\end{abstract}

\section{Introduction}
\input{sec/introduction.tex}

\section{Related Work}

\input{sec/related}

\section{Method}
\input{figs/model}

\input{sec/method}

\input{figs/expert_res}
\section{Experiments}
\input{sec/experiments}

\subsection{SUNCG}
\input{sec/ex_suncg}

\input{figs/scannet}

\subsection{Stereo Expert System}
\input{sec/ex_expert_system}

\input{figs/eth3d_res}

\input{figs/scannet_eth3d_table}
\subsection{ScanNet}
\input{sec/ex_scannet}

\subsection{ETH3D}
\input{sec/ex_eth3d}


\section{Conclusion}
\input{sec/conclusion}

\blfootnote{\small \textbf{Acknowledgements.} Denys Rozumnyi was supported by Czech Science Foundation grant GA18-05360S, CTU student grant SGS17/185/OHK3/3T/13 and ETH SSRF. Further support was received by the Intelligence Advanced Research Projects Activity (IARPA) via Department of Interior/ Interior Business Center (DOI/IBC) contract number D17PC00280. The U.S. Government is authorized to reproduce and distribute reprints for Governmental purposes notwithstanding any copyright annotation thereon. Disclaimer: The views and conclusions contained herein are those of the authors and should not be interpreted as necessarily representing the official policies or endorsements, either expressed or implied, of IARPA, DOI/IBC, or the U.S. Government. }

{\small
\bibliographystyle{ieee_fullname}
\bibliography{main}
}

\end{document}

%% file: sec/notation.tex
\newcommand{\depthMeas}{\textit{d}}
\newcommand{\gradMeas}{\textit{g}}
\newcommand{\nccMeas}{\textit{n}}

\newcommand{\boldparagraph}[1]{\vspace{0.2em}\noindent{\bf #1} }

\DeclareMathOperator*{\minimize}{minimize}

\newcommand{\nSet}[1]{\{#1\}}
\newcommand{\nProjSet}{\Pi}			      

\newcommand{\nVolDom}{\Omega}               

\newcommand{\nx}{\mathbf{x}}                
\newcommand{\nPrim}{u}                      
\newcommand{\nDual}{\xi}                    
\newcommand{\nData}{f}                      
\newcommand{\nConf}{c}                      
\newcommand{\nLagr}{\nu}                    
\newcommand{\nSens}{s}                      
\newcommand{\nSensSet}{\mathcal{S}}         

\newcommand{\nPDiter}{t}                    

\newcommand{\nRegW}{W}                      

\newcommand{\nLabelSet}{\mathcal{L}}        
\newcommand{\nLabel}{\ell}                  



%% file: sec/abstract.tex
Volumetric depth map fusion based on truncated signed distance functions has become a standard method and is used in many 3D reconstruction pipelines.
In this paper, we are generalizing this classic method in multiple ways:
1) Semantics: Semantic information enriches the scene representation and is incorporated into the fusion process.
2) Multi-Sensor: Depth information can originate from different sensors or algorithms with very different noise and outlier statistics which are considered during data fusion.
3) Scene denoising and completion: Sensors can fail to recover depth for certain materials and light conditions, or data is missing due to occlusions.
Our method denoises the geometry, closes holes and computes a watertight surface for every semantic class.
4) Learning: We propose a neural network reconstruction method that unifies all these properties within a single powerful framework.
Our method learns sensor or algorithm properties jointly with semantic depth fusion and scene completion and can also be used as an expert system, e.g. to unify the strengths of various photometric stereo algorithms.
Our approach is the first to unify all these properties. 
Experimental evaluations on both synthetic and real data sets demonstrate clear improvements.

%% file: sec/introduction.tex
Holistic 3D scene understanding is one of the central goals of computer vision research.
Tremendous progress has been made within the last decades to recover accurate 3D scene geometry with a variety of sensors \cite{Cui-et-al-CVPR-2010,Izadi-et-al-SIGGRAPH-2011,Mirdehghan-et-al-CVPR-2018} and image-based 3D reconstruction methods \cite{Furukawa-Ponce-TPAMI-2010,Vu-et-al-TPAMI-2012,Schoenberger-et-al-ECCV-2016}.
With the breakthrough in machine learning, algorithms that recover 3D geometry increasingly include semantic information \cite{Kim-et-al-ICCV-2013,Haene-et-al-CVPR-2013,Haene-et-al-TPAMI-2017,Blaha-et-al-CVPR-2016,Cherabier-et-al-3DV-2016,Kundu-et-al-ECCV-2014,Dai-et-al-CVPR-2017,Dai-et-al-CVPR-2018,Dai-Niessner-ECCV-2018,Cherabier-et-al-ECCV-2018,Tateno-et-al-CVPR-2017}
in order to improve the algorithm robustness, the accuracy of the 3D reconstruction and to provide a richer scene representation.
Many consumer products like smartphones, game consoles, augmented and virtual reality devices, as well as cars and household robots are equipped with an increasing amount of cameras and depth sensors.
Computer vision systems can highly benefit from this trend by leveraging multiple data sources and providing richer and more accurate results.
In this paper, we address the problem of multi-sensor depth map fusion for semantic 3D reconstruction.

Nowadays, depth can be estimated very robustly from multiple and even single RGB images \cite{Tateno-et-al-CVPR-2017}.
Nevertheless, depending on the camera, scene lighting, as well as the object and material properties, the noise statistics of computed depth maps can vary largely.
Moreover, popular depth sensors like the Kinect have varying noise statistics \cite{Zeisl-Pollefeys-ICRA-2016} depending on the depth value and the pixel distance to the image center. They also have trouble recovering depth on object edges as well as on light reflecting or absorbing surfaces, but perform well on low-textured surfaces and within short depth ranges.
In contrast, image-based stereo methods usually perform well on object edges and across a wide depth range, but fail on low-textured surfaces and have comparably high noise and outlier rates.
%
%
\input{figs/teaser.tex}

While traditional methods have tried to model these effects, they usually impose strong  assumptions about noise distribution, or they require tedious calibration to estimate all parameters \cite{Zeisl-Pollefeys-ICRA-2016}.
In contrast, we leverage the strength of machine learning techniques to extract sensor properties and scene parameters automatically from training data and use them in form of confidence values for a more accurate semantic depth map fusion.
Fig.~\ref{fig:teaser} shows example output of our method.
In sum, we make the following \textbf{contributions}:
\begin{itemize}[topsep=1pt,leftmargin=*]
\setlength\itemsep{-1mm}
\item We propose the first method to unify semantic 3D reconstruction, scene completion and multi-sensor data fusion into a single machine-learning-based framework. Our approach uses only few model parameters and thus needs only small amounts of training data to generalize well.
\item Our method analyses the sensor output and learns depth sensor-specific noise and outlier statistics which are considered when estimating confidence values for the TSDF fusion. 
For the case that the depth source is an algorithm we feed in both information about the depth output and information about the input patches such that out network is better able to learn when the algorithm typically fails.
\item Besides the multi-sensor data fusion, our approach can also be used as an expert system for multi-algorithm depth fusion in which the outputs of various stereo methods are fused to reach a better reconstruction accuracy.
\end{itemize}

%% file: figs/teaser.tex
\begin{figure}[t!]
  \small
  \centering
  \setlength{\tabcolsep}{1mm}
  \newcommand{\sz}{0.36}
  \begin{tabular}{ccc}
    \includegraphics[height=\sz\columnwidth]{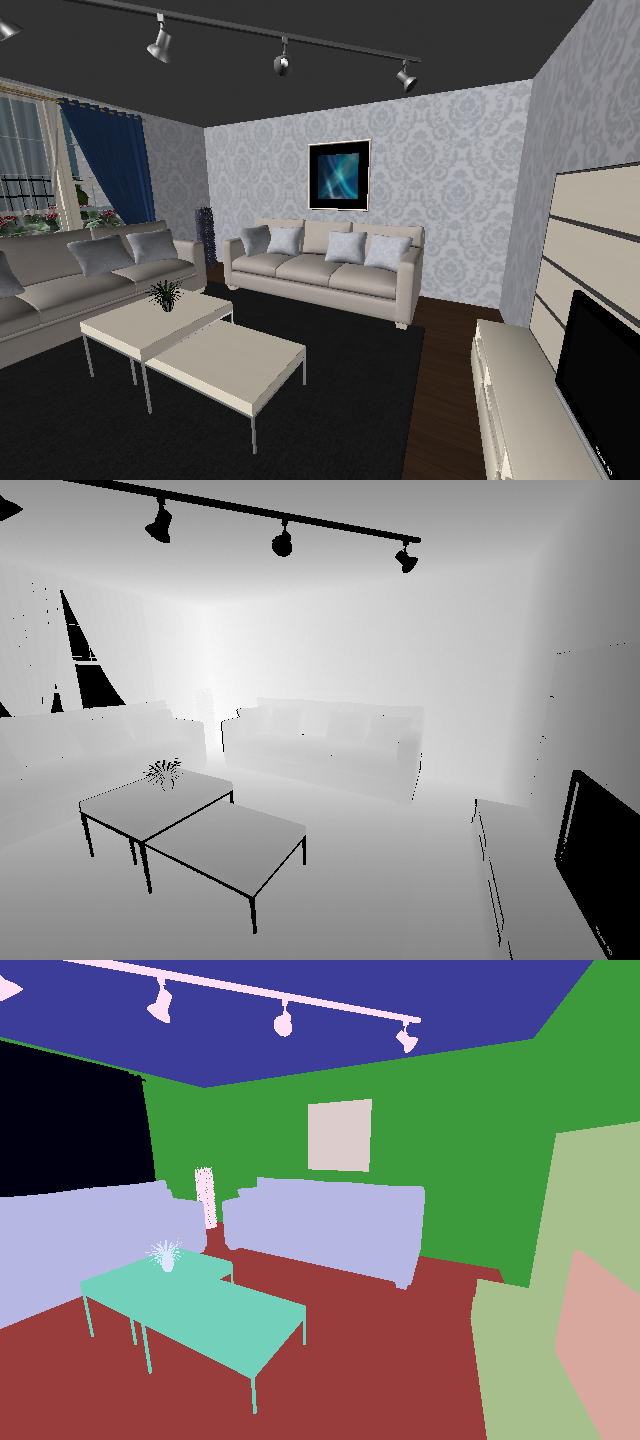} &
    \includegraphics[height=\sz\columnwidth]{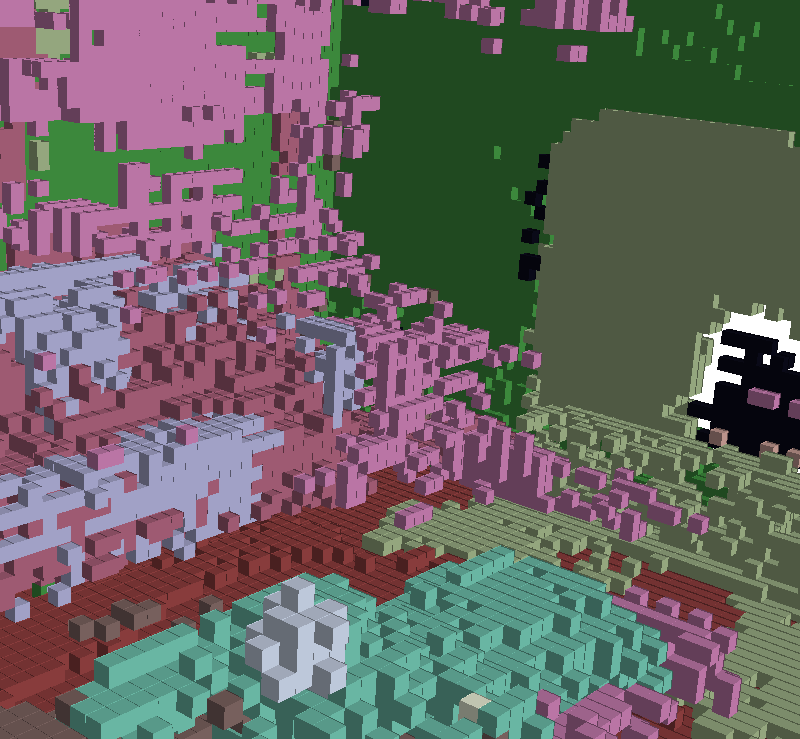} &
    \includegraphics[height=\sz\columnwidth]{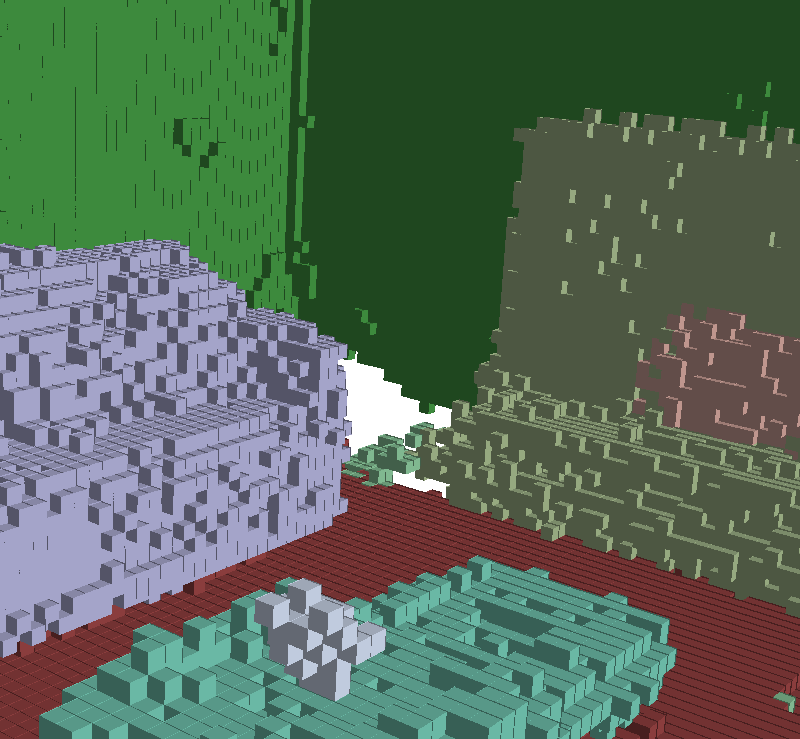}\\
    Inputs & \cite{Cherabier-et-al-ECCV-2018} (Std. TSDF Fusion) & Learned Fusion (ours) \\[0.1cm]
  \end{tabular}
  \caption{\textbf{Depth map fusion of Kinect and photometric stereo.}
	Our fusion approach learns sensor noise and outlier statistics and accounts them via confidence weights in the fusion process. This yields more accurate and more complete semantic reconstructions.
  }
  \label{fig:teaser}
\end{figure}

%% file: sec/related.tex
\boldparagraph{Volumetric Depth Fusion.}
In their pioneering work, Curless and Levoy~\cite{Curless-Levoy-SIGGRAPH-1996} proposed a simple and effective method to fuse depth maps from multiple views by averaging \textit{truncated signed distance functions} (TSDFs) within a regular voxel grid.
With the broad availability of low-cost depth sensors like the MS Kinect, this method became very popular with influential works like KinectFusion~\cite{Izadi-et-al-SIGGRAPH-2011} and its numerous extensions, like voxel hashing~\cite{Niessner-et-al-SIGGRAPH-2013} or voxel octrees~\cite{Steinbruecker-et-al-ICCV-2013}.
This depth fusion method has become standard for SLAM frameworks like InfiniTAM~\cite{Kaehler-et-al-TVCG-2015} and was further generalized to account for drift and calibration errors, e.g. ElasticFusion~\cite{Whelan-et-al-IJRR-2016}, BundleFusion~\cite{Dai-et-et-al-SIGGRAPH-2017}, but also for 3D reconstruction frameworks~\cite{Zach-et-al-ICCV-2007,Kolev-et-al-IJCV-2009,Haene-et-al-CVPR-2013,Haene-et-al-TPAMI-2017,Dai-Niessner-ECCV-2018,Cherabier-et-al-ECCV-2018}.

All these methods have in common that TSDF fusion is performed via simple uniformly weighted averaging.
Hence these methods do not account for the fact that depth measurements may exhibit different noise and outlier rates.
This has been tackled by probabilistic fusion methods.

\boldparagraph{Probabilistic Depth Fusion.}
Probabilistic approaches explicitly model sensor noise, typically with a Gaussian distribution.
A very simple approach with only 2.5D output and a Gaussian noise assumption can be found in \cite{Duan-et-al-ICPR-2012}.
A point-based fusion approach is proposed in \cite{Keller-et-al-3DV-2013}.
Instead of a voxel grid, the fusion updates are directly performed on a point cloud.
This has been extended to anisotropic point-based fusion in \cite{Lefloch-et-al-FUSION-2015} to account for different noise levels when a surface is observed from a different viewing angle.
For a fixed-topology the mesh-based fusion approach by \cite{Zienkiewicz-et-al-3DV-2016} fuses depth information over various mesh resolutions.
A more complex probabilistic fusion method is proposed in \cite{Woodford-Vogiatzis-ECCV-2012} which includes long range visibility constraint in their online fusion method.
A similar model with long-range ray-based visibility constraints was used in \cite{Ulusoy-et-al-3DV-2015,Ulusoy-et-al-CVPR-2016}, although these methods are not real-time capable.
Recently, PSDF Fusion~\cite{Dong-et-al-ECCV-2018} demonstrated a combination of probabilistic modeling and a TSDF scene representation.
However, they also assume a Gaussian error distribution of the input depth values.
Overall, probabilistic approaches handle noise and outliers better than traditional TSDF fusion methods.
Nevertheless, the majority of these methods impose strong assumptions about the sensor error distributions to define the prior model.
The first method that implicitly learns an unknown error distribution during the fusion is OctNetFusion by Riegler~\etal \cite{Riegler-et-al-3DV-2017}.
They jointly learn the splitting of the octree scene representation, but multiple sensors or semantic information are not considered.

\boldparagraph{Multi-Sensor Data Fusion.}
Early approaches like Zhu~\etal\cite{Zhu-et-al-CVPR-2008} fuse time-of-flight depth and stereo, but only for a 2.5D depth map.
Kim~\etal~\cite{Kim-et-al-ICCVW-2009} fuse the same sensor combination with 3D via a probabilistic framework on a voxel grid. Work by~\cite{Chiu-et-al-BMVC-2011} strives for low-level data fusion to improve the Kinect output with stereo correspondences.
As an extension of \cite{Duan-et-al-ICPR-2012}, Duan~\etal\cite{Duan-et-al-ROBIO-2012} use a probabilistic approach for the fusion of Kinect and Stereo in real-time.
None of the current multi-sensor depth fusion networks is able to incorporate semantic information and their generalization is usually non-trivial.

\boldparagraph{3D Reconstruction with Confidences.}
A wide range of 3D reconstruction approaches estimate confidence values for depth hypotheses which are then later used for adaptive fusion.
All these approaches typically use either handcrafted confidence weights \cite{Fuhrmann-Goesele-TOG-2014,Ummenhofer-Brox-IJCV-2017,Kuhn-et-al-IJCV-2017} rather than learning them intrinsically from data or they learn only 2D score map without learning their 3D fusion \cite{Pogg-Mattoccia-BMVC-2016,Tosi-et-al-BMVC-2017,Tosi-et-al-ECCV-2018,Weerasekera-et-al-ICRA-2018}.

\boldparagraph{Semantic 3D Reconstruction and Scene Completion.}
Joint semantic label estimation and 3D geometry has been proposed with traditional energy-based methods to estimate depth maps \cite{Ladicky-et-al-IJCV-2012} or dense volumetric 3D \cite{Kim-et-al-ICCV-2013,Haene-et-al-CVPR-2013,Haene-et-al-TPAMI-2017,Blaha-et-al-CVPR-2016,Cherabier-et-al-3DV-2016,Kundu-et-al-ECCV-2014}. 
Machine learning-based approaches have pushed the state of the art in reconstructing and completing 3D scenes
%
\cite{Dai-et-al-CVPR-2017,Dai-et-al-CVPR-2018,Dai-Niessner-ECCV-2018,Cherabier-et-al-ECCV-2018}.
These methods are not real-time capable, but real-time fusion of CNN-based single-image depth and semantics has recently been presented in CNN-SLAM~\cite{Tateno-et-al-CVPR-2017}.

So far none of the semantic 3D reconstruction approaches is able to properly handle multiple sensors with different noise characteristics and their extension is not straightforward.
Our goal is a general framework which unifies all the previously discussed properties within a learning-based method.

%% file: figs/model.tex
\begin{figure*}[ht!] 
  \centering
  \includegraphics[width=\textwidth]{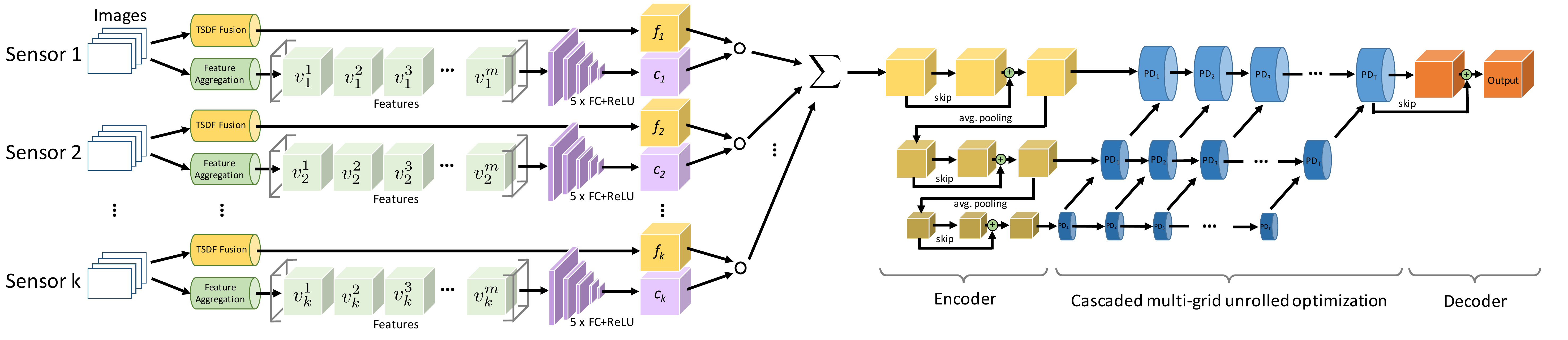}
  \vspace{-0.5cm}
  \caption{\textbf{Network architecture.} Our network consists of two connected networks which are jointly trained.  \textbf{Left: Sensor Confidence Network} which aggregates voxel-wise confidence values for each sensor. First we fuse semantic TSDFs (yellow) and aggregate features (green) from the input depth maps and images. Then a small fully connected network with ReLU activations processes the features and predicts a confidence weight (magenta).  \textbf{Right: Semantic 3D Reconstruction Network} which performs 3D reconstruction, denoising and scene completion. This network consists of special layers (blue) which minimize an energy that denoises and completes the scene within a multi-grid setting and finally outputs semantically labeled occupancy grid (red). The right network part corresponds to the one in \cite{Cherabier-et-al-ECCV-2018}.}
  \label{fig:model_overview}
\end{figure*}

%% file: sec/method.tex
For performing semantic 3D reconstruction, our method requires as input a set of RGB-D images and their corresponding 2D semantic segmentations as shown in Fig.~\ref{fig:teaser}. 
The semantic segmentations can be fused into the TSDF representation of the scene, using~\cite{Haene-et-al-CVPR-2013}.
In the following, we describe how we can robustly produce an accurate TSDF by fusing measurements from multiple depth sensors.

\boldparagraph{Key idea.}
We consider multiple depth sensors which produce a set of depth maps by scanning a scene. 
The most common approach to data fusion consists in fusing all the depth maps, regardless of the sensor that produced them, into a TSDF representation of the scene.
However, this does not reflect the specific noise and outliers statistics of each measurement.
We propose to overcome this issue by learning a confidence estimator for every sensor that weights the measurements before fusing them.
For each sensor, we can produce a TSDF representation of the scene by fusing the corresponding depth maps.
Our method learns to estimate confidence values for every voxel in TSDF, such that the accuracy of the semantic 3D reconstruction is maximized.

We propose an end-to-end trainable neural network architecture which can be roughly separated into two parts: a \textit{sensor confidence network} which predicts a confidence value for each sensor measurement, and a \textit{semantic 3D reconstruction network} which takes all aggregated noisy measurements and corresponding confidences and performs semantic 3D reconstruction.

The overall network structure is depicted in Fig.~\ref{fig:model_overview} and the individual network parts are detailed in the following subsections.

\subsection{Sensor Confidence Network}

\boldparagraph{Weighted TSDF Fusion.}
A sensor $i$ produces a set of depth maps that can be fused into a TSDF $\nData_i$, following \cite{Curless-Levoy-SIGGRAPH-1996}.
We learn to estimate corresponding confidence maps $c_i$, where for every voxel $\nx$, $c_i(\nx)$ is the confidence for the measurement $\nData_i(\nx)$.
The fusion of all the sensor measurements is then computed via a point-wise weighted average:
\begin{equation}
  \nData(\nx) = \frac{\sum c_i(\nx) \nData_i(\nx)}{\sum c_i(\nx)}
\end{equation}
%


%

\boldparagraph{Goal.}
The purpose of the confidence weight learning for multi-sensor TSDF fusion is twofold:
\textbf{1) Intra-Sensor Weighting:} The network captures the noise and outlier statistics among measurements thus producing a spatially varying confidence map, \eg points that are mostly observed from a far distance can get a lower confidence than those mainly observed from a closer distance.
%
\textbf{2) Inter-Sensor Weighting:}
The network analyses the noise and outlier statistics among different sensors in order to weight them against each other.
In this regard the network also accounts for normalization which is important if there are different amounts of data available from different sensors.
This avoids for instance a bias towards a sensor with a higher frame rate.

\boldparagraph{Feature extraction.}
%
%
%
We aggregate features from the input data which we believe will help the network to estimate a reliable confidence value.
Ideally, we could feed all input data into our confidence network and the network could identify important features for the confidence estimation on its own, but the amount of input data for the scenes we consider in this paper renders this infeasible.
Therefore, our selected feature set is certainly not exhaustive and there might be other useful features or better feature combinations.
However, we found all of them improving the reconstruction results.
For each sensor $k$ and each voxel, we extract the following $m=13$ features $\nSet{v_k^l}_{l=1}^m$:
\begin{itemize}[topsep=2pt,leftmargin=*]
\setlength\itemsep{0mm}
  \item \textbf{Average 3$\times$3 patches in depth image} (9 values): Analyzing neighboring depth values helps to identify outliers in the depth map (Fig. \ref{fig:feat_extrac}).
  \item \textbf{Mean and standard deviation of image gradient norm on 3$\times$3 patches} (2 values): Especially for stereo methods the average gradient norm of a patch indicates how much gradient information is contained in the patch. Homogeneously colored patches should lead to low confidence values. 
  \item \textbf{Mean and standard deviation of normalized cross correlation (NCC) of stereo 5$\times$5 patches (in case of stereo algorithms)} (2 values): NCC is an established measure for estimating patch similarity for stereo methods. If the patches do not match well, or there is a high variance of NCC values among patches voting for the same point, then the confidence value should be reduced.
\end{itemize}
This set of features is then processed for each voxel individually by a small neural network which estimates a confidence weight for a single voxel (magenta in Fig.~\ref{fig:model_overview}).

\input{figs/feature_extrac.tex}

\boldparagraph{Confidence Network Architecture.}
%
%
%
The small confidence estimation networks have identical structure for each sensor and identical weights for each voxel of a sensor. 
They consist of 5 fully connected neural layers with ReLU activations and with a decreasing number of neurons $\nSet{100, 50, 20, 10, 1}$. 
The last layer is initialized with biases equal to one such that the initial confidence values are equal for each sensor. 
The remaining weights are initialized randomly.
The output of the confidence networks are then aggregated into a single TDSF volume which serves as input for a semantic 3D reconstruction network.

\subsection{Semantic 3D Reconstruction Network}
%
Our approach learns in an end-to-end fashion how to jointly perform data fusion and semantic 3D reconstruction.
The data fusion should facilitate the semantic 3D reconstruction by providing additional and more complete information about the scene.
To perform the reconstruction, we use the architecture introduced in \cite{Cherabier-et-al-ECCV-2018} which leverages the benefits of neural networks and variational methods.
The fundamental principle of the method is to compute a consistent voxel labeling from noisy and incomplete depth such that semantic voxel transitions are statistically similar to the transitions previously seen in the training data.
For instance, a bed should be standing on the ground, with vertical transitions to the ground below and the free space above, while a wall should have a horizontal transitions to free space.

The motivations are the following:
\begin{itemize}[topsep=2pt,leftmargin=*]
\setlength\itemsep{0mm}
\item The architecture, which relies on the principles of total variation segmentation and inpainting, contains very few parameters to learn due to weight sharing.
Due to the few parameters the network does not need much training data which is beneficial since only few and small real data sets are available for training.
\item The compact architecture allows to easily extend the network to estimate further parameters for the data fusion and still allows to process larger scenes with more then 15M voxels.
\item The energy formulation allows us to incorporate an arbitrary number of sensors into the 3D reconstruction method, which is more difficult with standard feed-forward architectures.
\end{itemize}

\input{figs/suncg_res}

\boldparagraph{Variational method.} 
We briefly describe the working principles of the reconstruction network.
More details can be found in \cite{Cherabier-et-al-ECCV-2018}.
At its core, the network minimizes an energy such that the solution corresponds to a scene with label transition statistics that match the training data. 
We define $\nVolDom$ the voxel grid, and write the energy as:
\begin{align}
  &\minimize_\nPrim \quad
  \int_\nVolDom \Big( \|\nRegW \nPrim\|_2 + 
  \sum_{\nSens\in\nSensSet}(\nConf_\nSens\!\circ\!\nData_\nSens)\; \nPrim \Big) \;d\nx
  \label{eq:sem3D_energy} \\
  &\text{subject to} \quad \forall \nx\!\in\!\nVolDom: \thinspace 
  \sum_{\nLabel\in\nLabelSet} \nPrim_\nLabel\left(\nx\right) \!=\! 1
  \nonumber
\end{align}
In Eq.~\eqref{eq:sem3D_energy}, $\nPrim$ is the voxel labeling we optimize for, defined such that $\nPrim_\nLabel(\nx)\in [0,1]$ is the probability that label $\nLabel$ is given to voxel $\nx$.
The operator $\circ$ denotes element-wise multiplication (Hadamard product).
The operator $\nRegW$ is a regularizer that enforces the labeling to respect certain conditions on the semantic transitions (\eg the bed stands on the ground).
During training, $\nRegW$ is learned to capture typical scene statistics.
This can be implemented as a convolution which locally compares voxels to their neighborhood, thus verifying the semantic transitions.

The energy \eqref{eq:sem3D_energy} is numerically minimized with a first-order algorithm \cite{Chambolle-Pock-JMIV-2011}.
To this end, dual variables $\nDual,\nLagr$ are introduced to account for the non-differentiability and the constraint in Eq.~\eqref{eq:sem3D_energy}, leading to the following equivalent discretized saddle point energy 
\begin{equation}
  \minimize_\nPrim \max_{\|\nDual\|_\infty \leq 1} 
    \langle \nRegW\nPrim, \nDual \rangle + 
    \sum_{\nSens\in\nSensSet} \langle \nConf_\nSens\!\circ\!\nData_\nSens, \nPrim \rangle + 
    \nLagr\Big( 1\!-\! \sum_{\nLabel\in\nLabelSet}\!\nPrim_\nLabel \Big)
  \label{eq:pd_energy}
\end{equation}
The numerical minimization iterations are unrolled and each layer of our network (blue cylinders in Fig.~\ref{fig:model_overview}) performs the following updates to minimize energy \eqref{eq:pd_energy}.
The inputs and outputs of each layer are shown on the left.
\begin{equation}
  \substack{
    \xRightarrow[\enskip\;\nPrim^\nPDiter,\bar{\nPrim}^\nPDiter\enskip\;]
                {\nLagr^\nPDiter,\nDual^\nPDiter}\\[3em]
    \xLeftarrow[\nPrim^{\nPDiter+1},\bar{\nPrim}^{\nPDiter+1}]
               {\nLagr^{\nPDiter+1},\nDual^{\nPDiter+1}}
  }
  \begin{cases}
    \nLagr^{\nPDiter+1} &= \nLagr^{\nPDiter} \!+\! \sigma\left(
      \sum_{\nLabel} \bar{\nPrim}_\nLabel^{\nPDiter} - 1\right) 
      \\ 
    \nDual^{\nPDiter+1} &= \nProjSet_{\|\cdot\|\leq 1}
      \left[ \nDual^\nPDiter + \sigma\nRegW\bar{\nPrim}^\nPDiter \right]
      \\ 
    \nPrim^{\nPDiter+1} &= \nProjSet_{\left[0, 1\right]} 
      \Big[\nPrim^\nPDiter \!-\! 
      \tau(\nRegW^*\nDual^{\nPDiter+1} + \\
      & \qquad\qquad \nLagr^{\nPDiter+1} \!+\!
      \sum_{\nSens\in\nSensSet}\nConf_\nSens\!\circ\!\nData_\nSens) \Big]
    \\
    \bar{\nPrim}^{\nPDiter+1} &= 
      2\nPrim^{\nPDiter+1} - \nPrim^{\nPDiter}
  \end{cases}
\label{eq:pd_updates}
\end{equation}
For better readability these steps show the single resolution variant. For the multi-grid version the update steps for $\nDual$ and $\nPrim$ change slightly (please see \cite{Cherabier-et-al-ECCV-2018} for more details).

%% file: figs/feature_extrac.tex
\begin{figure}[t!]
  \small
  \centering
  \newcommand{\sz}{0.65}
    \includegraphics[height=\sz\columnwidth]{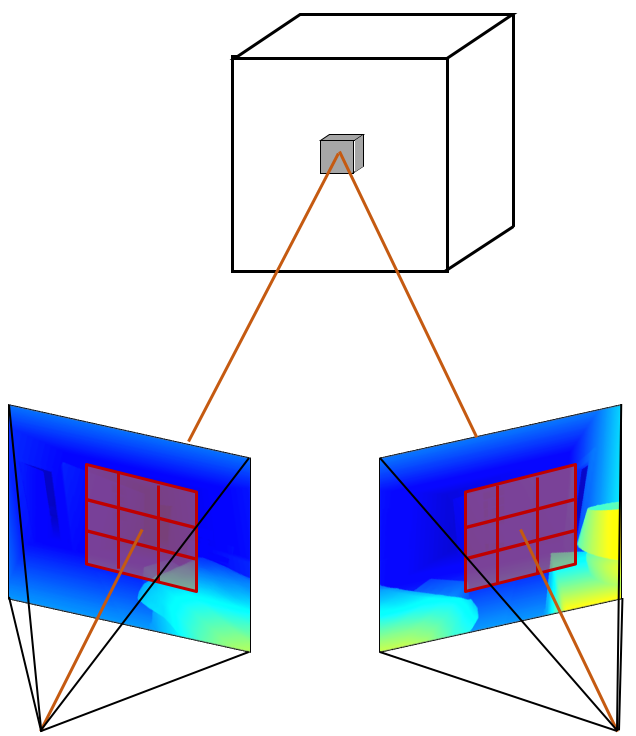}
  \vspace{0.1cm}
  \caption{\textbf{Voxel-wise feature extraction.}
	Extraction of the average depth on a 3x3 patch. The voxel is back-projected onto the depth maps, and the patches are centered around the back-projections (represented in red). This approach is identical for the extraction of the gradient and patch similarity features.
  }
  \label{fig:feat_extrac}
  \vspace{-0.1cm}
\end{figure}

%% file: figs/suncg_res.tex
\newcommand{\scannetFigHg}{0.22}
\begin{figure*} 
\centering
\begin{tabular}{cccc}
\includegraphics[width=\scannetFigHg\textwidth]{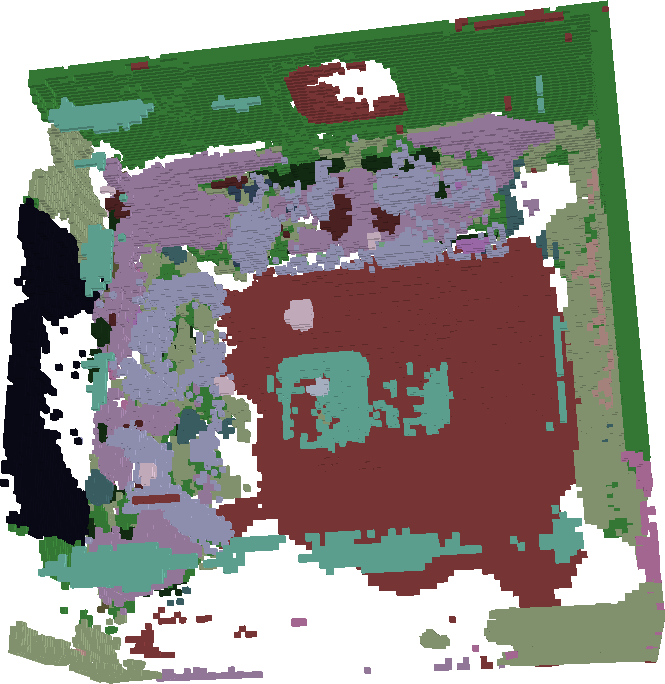} &
\includegraphics[width=\scannetFigHg\textwidth]{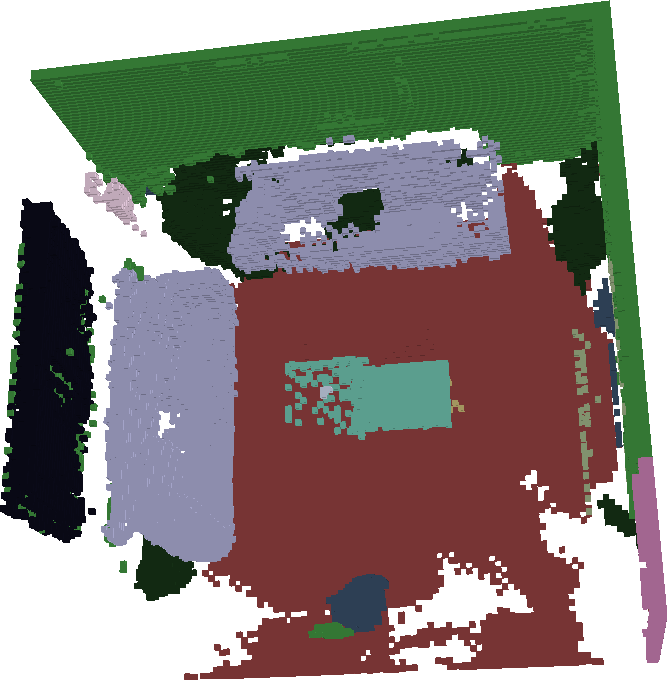} &
\includegraphics[width=\scannetFigHg\textwidth]{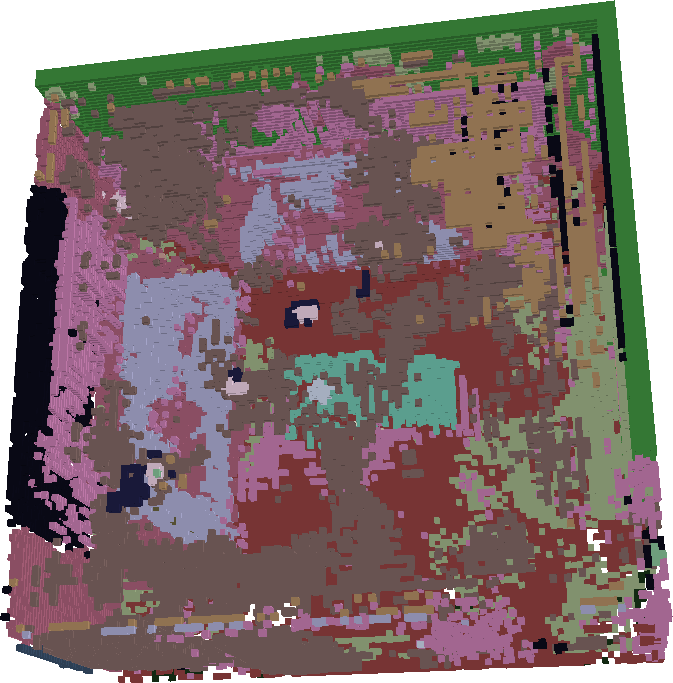} &
\includegraphics[width=\scannetFigHg\textwidth]{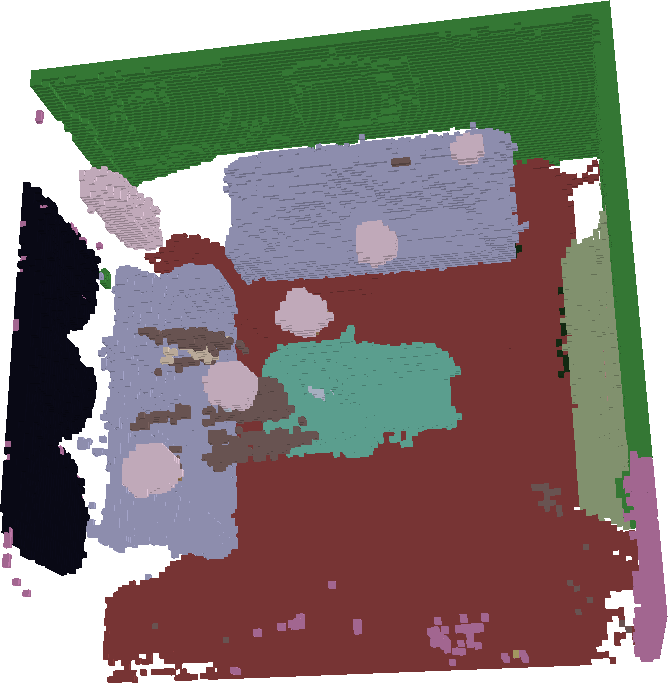} \\

Stereo (S) full: 0.71 &                                    
Kinect (K) full: 0.77 & 	                         
S half + K half: 0.76  &    
S half + K half (\depthMeas): 0.77  \\

\includegraphics[width=\scannetFigHg\textwidth]{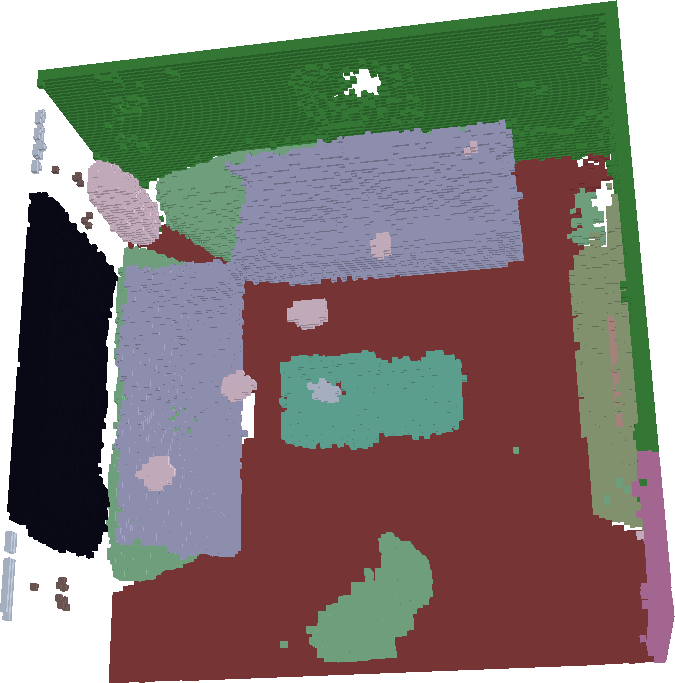} &
\includegraphics[width=\scannetFigHg\textwidth]{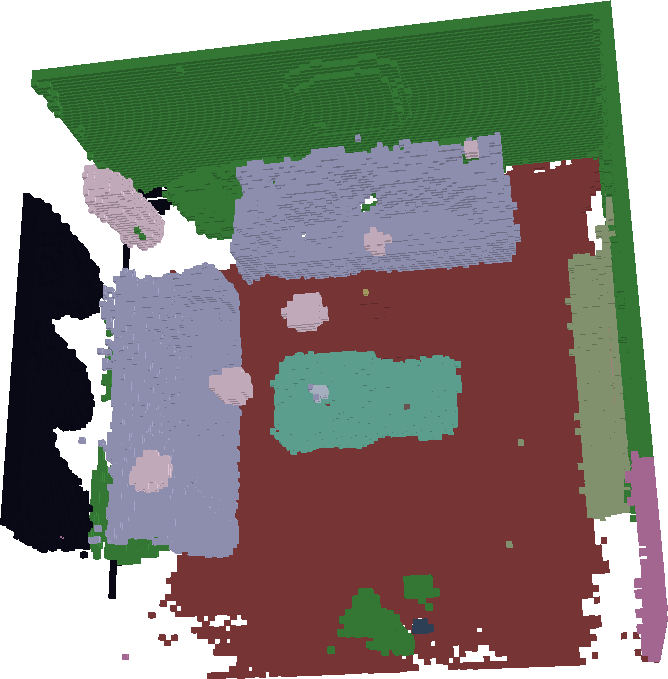} &
\includegraphics[width=\scannetFigHg\textwidth]{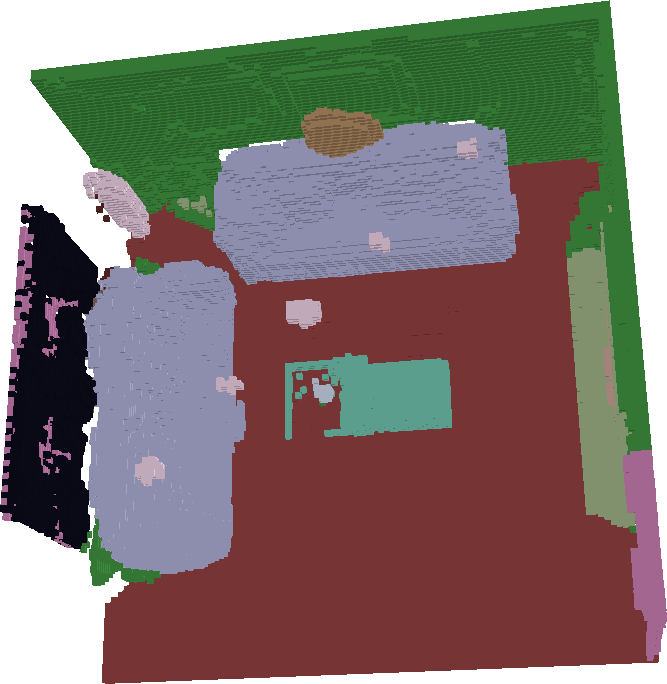} &
\includegraphics[width=\scannetFigHg\textwidth]{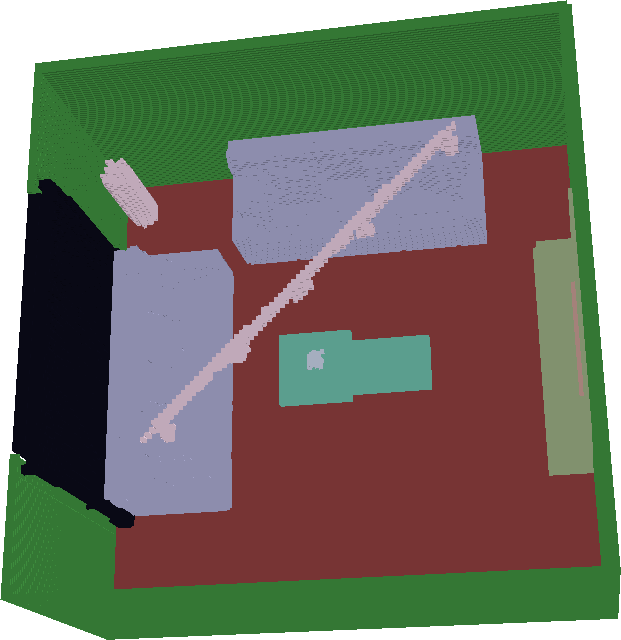} \\

S half + K half (\depthMeas, \gradMeas): 0.78 &
S full + K full (\depthMeas, \gradMeas): 0.786  & 	                     
Perfect sensor full: 0.794  &
Ground Truth \\[4pt]
\end{tabular}
\caption{\textbf{Ablation study on SUNCG.} For each step we report average semantic accuracy on the whole dataset. Average depth patches for confidence estimation are marked by \depthMeas, and gradient mean and standard deviation by \gradMeas. "Full" means that all views were used, whereas "half" means that views were split in two parts. Former refers to noise canceling, and latter to scene completion.}\label{fig:suncg}
\end{figure*}

%% file: figs/expert_res.tex
\newcommand{\expertFigHg}{0.22}
\begin{figure*} 
\centering
\begin{tabular}{cccc}

\includegraphics[width=\expertFigHg\textwidth]{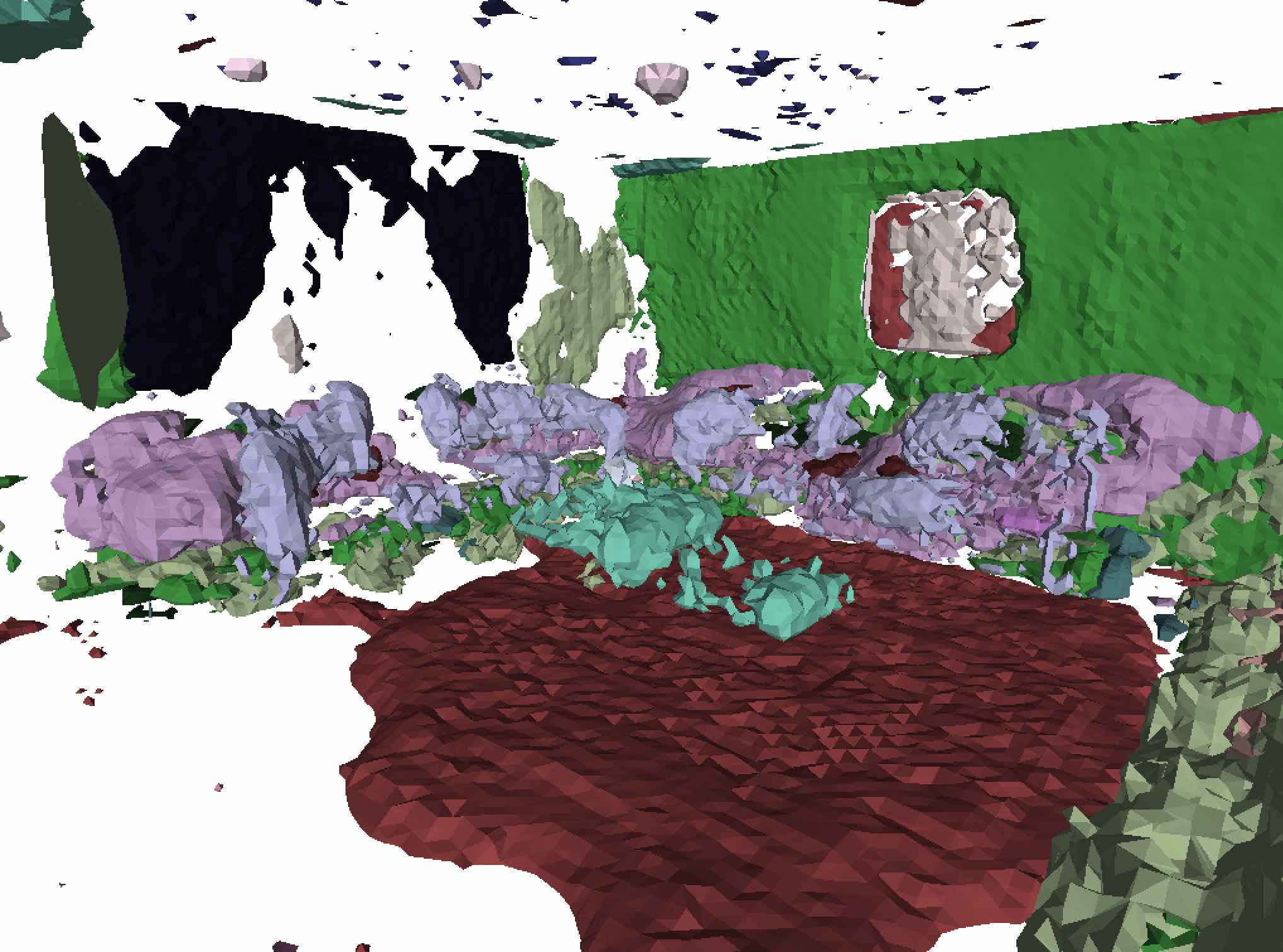} &
\includegraphics[width=\expertFigHg\textwidth]{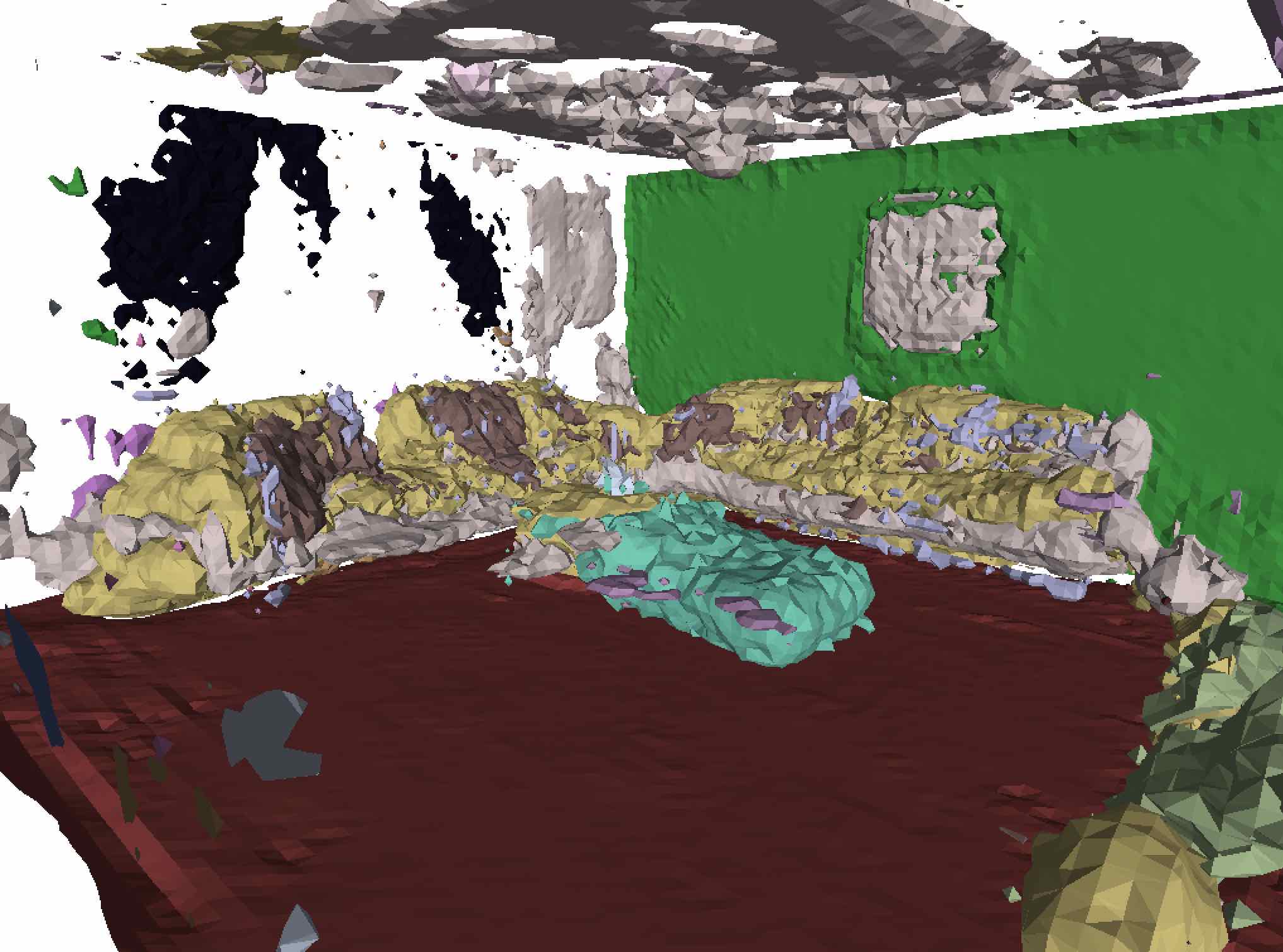} &
\includegraphics[width=\expertFigHg\textwidth]{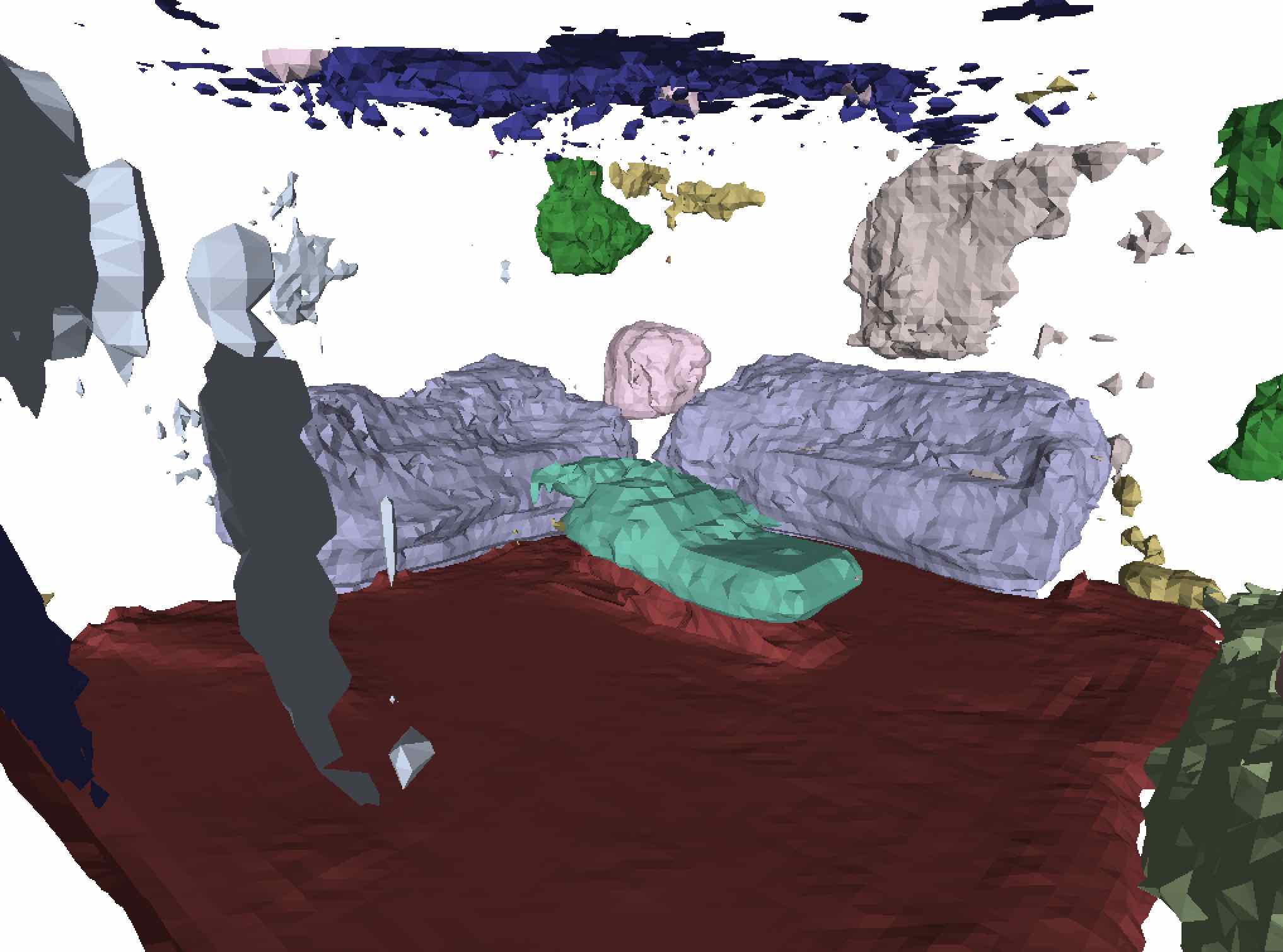} &
\includegraphics[width=\expertFigHg\textwidth]{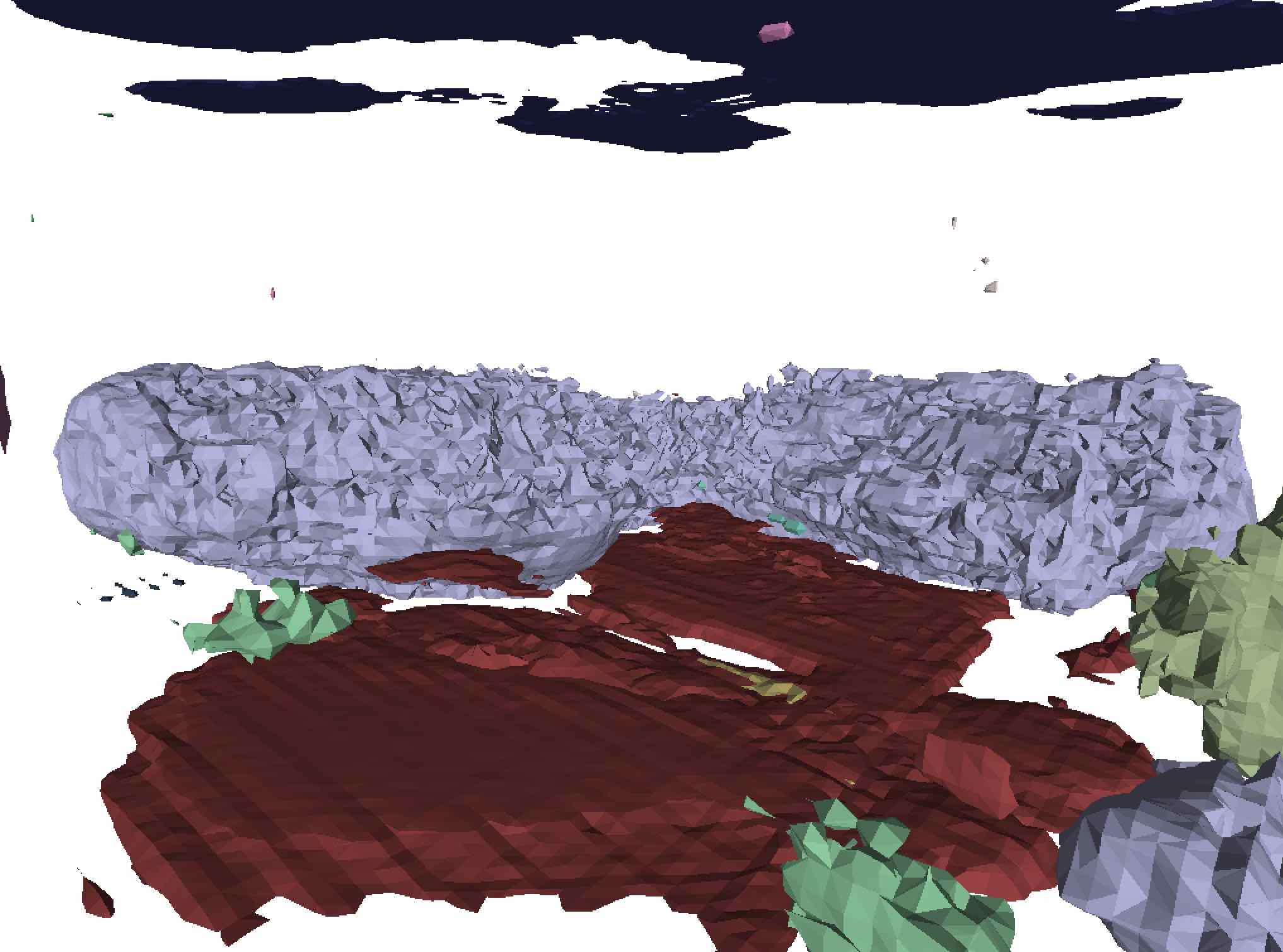} \\

A) SGBM~\cite{sgbm}: 0.71  &                                                
B) BM~\cite{opencv_library}: 0.71  &	      
C) PSMNet~\cite{psmnet}: 0.69  &		      
D) FCRN monocular~\cite{fcrn}: 0.44  \\

\includegraphics[width=\expertFigHg\textwidth]{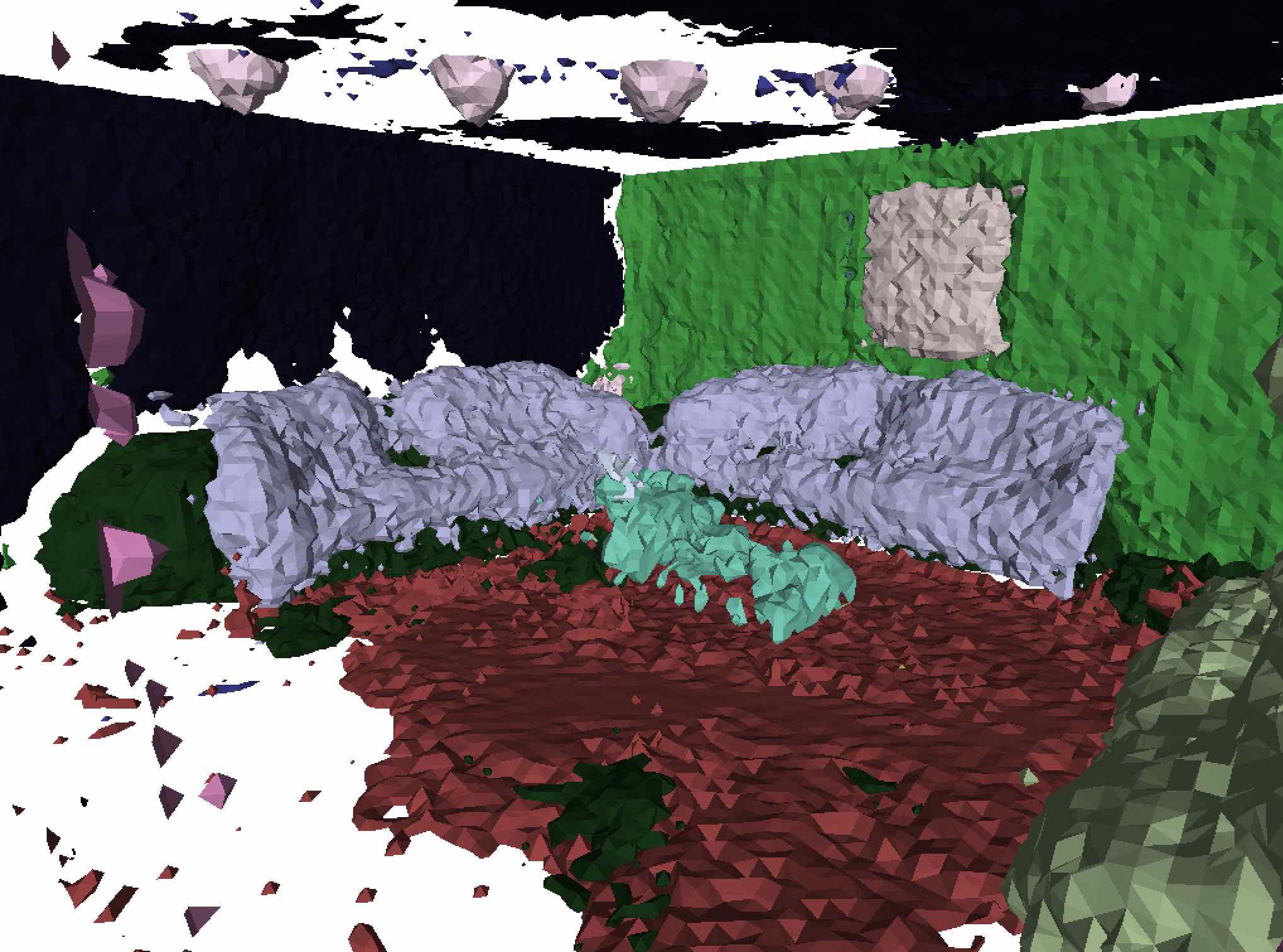} &
\includegraphics[width=\expertFigHg\textwidth]{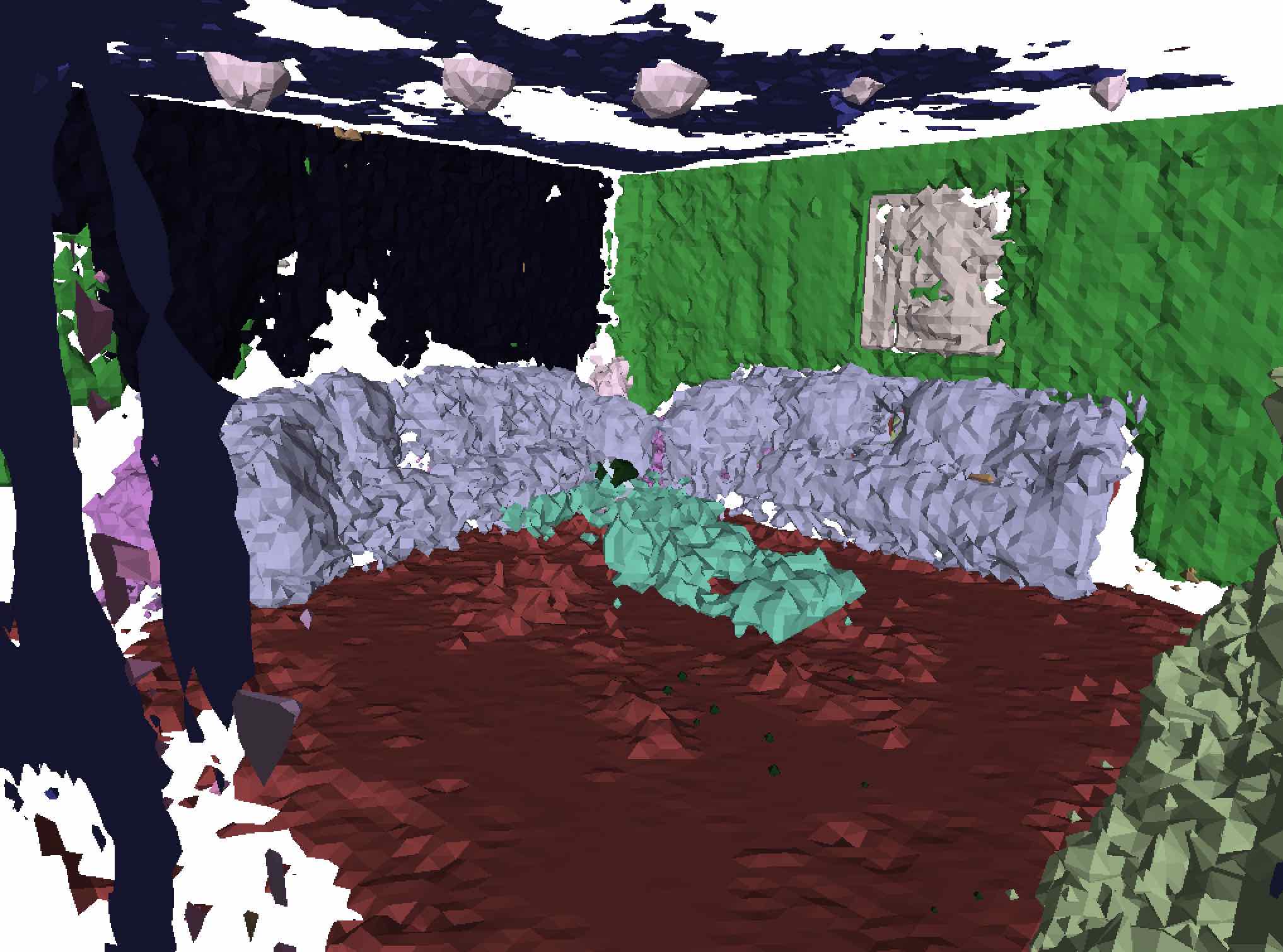} &
\includegraphics[width=\expertFigHg\textwidth]{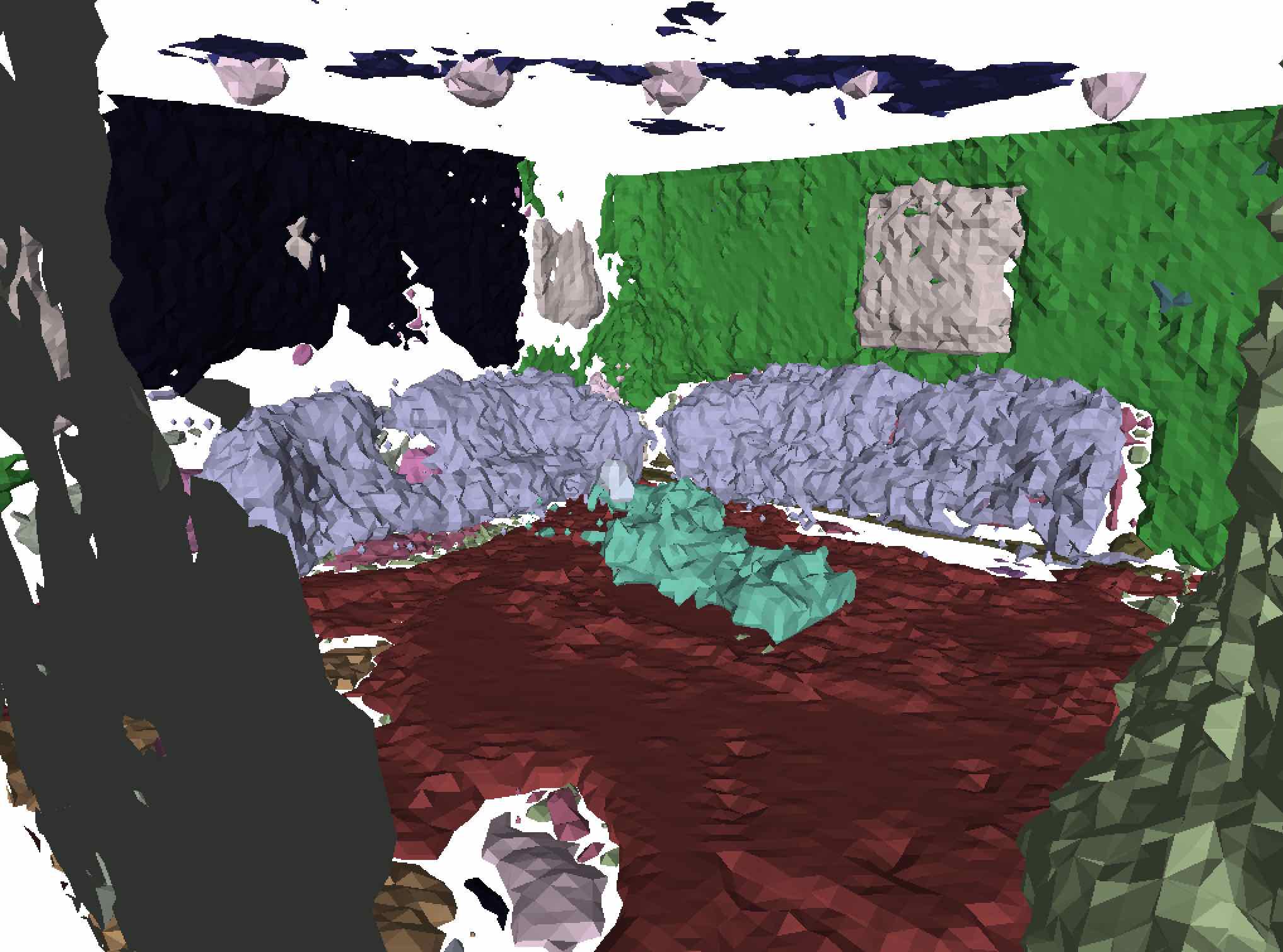} &
\includegraphics[width=\expertFigHg\textwidth]{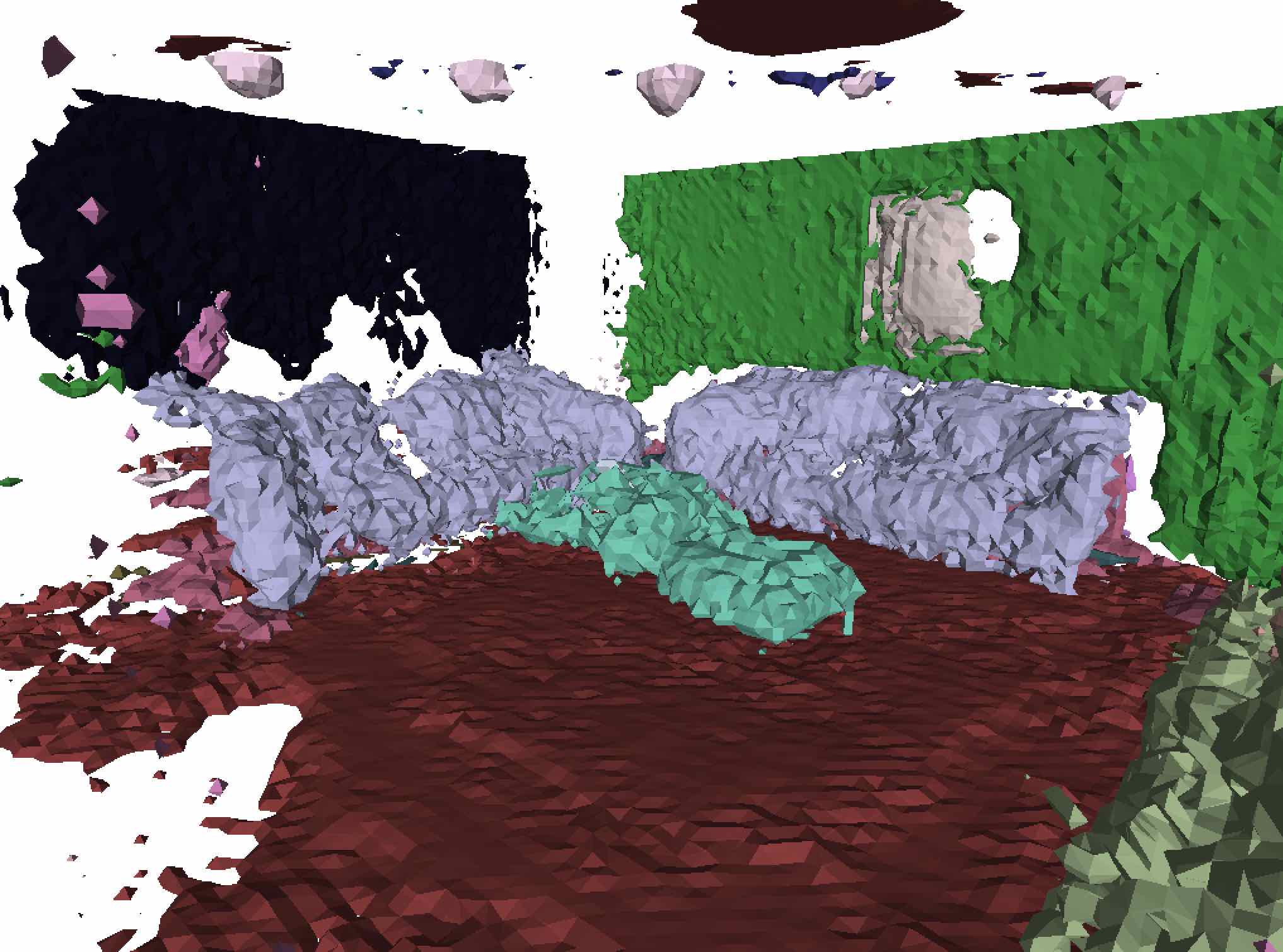} \\

A+B+C (\depthMeas): 0.72 &	                         
A+B+C (\depthMeas, \gradMeas): 0.725  &
A+B+C (\depthMeas, \gradMeas, \nccMeas): 0.735  & 	 
A+B+C+D (\depthMeas, \gradMeas): 0.73  \\[4pt]	  
\end{tabular}
\caption{\textbf{Close-up views for the expert system experiment with learned fusion of 4 stereo algorithms}: Block Matching (BM)~\cite{opencv_library}, Semi-Global Block Matching (SGBM)~\cite{sgbm}, PSMNet~\cite{psmnet}, FCRN monocular~\cite{fcrn}. The top row contains TSDF fusion results from each of the stereo algorithms separately, whereas the bottom row provides results of different algorithm combination as well as different types of input used for confidence estimation. Average depth patches for confidence estimation are denoted by \depthMeas, gradient mean and standard deviation by \gradMeas, and average normalized cross correlation of stereo patches by \nccMeas. FCRN is a monocular method and thus \nccMeas~ cannot be computed. For each method we report average semantic accuracy on the whole dataset.}\label{fig:expert}
\end{figure*}

%% file: sec/experiments.tex
\boldparagraph{Setup and Implementation.}
The entire framework has been implemented in Python/Tensorflow and runs on a computer with E5-2630 processor and an NVidia GTX 1080 Ti GPU running a recent Linux distribution.
The network training was done with the ADAM optimizer~\cite{adamopt}, with learning rate $0.0001$ and batch size $4$. 
All training samples were random crops of the input data of dimension $(24,24,24)$.
Then every crop was randomly rotated around the $z$-axis and randomly flipped along $x$ and $y$ axes. 
The network was trained for 1000 epochs, which was enough to converge for all datasets. 
One epoch iterated once over all scenes.
The number of hierarchical levels was set to 3 and number of unrolled optimization iterations to 50, as in~\cite{Cherabier-et-al-ECCV-2018}. 
On average training required about 3 hours for 1000 epochs. Inference of one scene takes 3 to 5 minutes on GPU.

\boldparagraph{Datasets.}
The experiments were done on three datasets: SUNCG~\cite{suncg}, ScanNet~\cite{scannet} and ETH3D~\cite{eth3d}. 
For every dataset and experiment we measure semantic and free-space accuracy. 
Semantic accuracy (SA) is defined as a ratio of occupied voxels (\ie non free space) for which the particular semantic label was estimated correctly, divided by the total number of occupied voxels.
Similarly, the free-space accuracy (FA) is a ratio of voxels, for which the unique free-space label was estimated correctly, divided by the number of free-space voxels. 
Splitting accuracy into two parts helps to account for domination of free-space voxels in all scenes.  
Then, the loss function is defined as categorical cross entropy, separately computed for semantic voxels as $L_s$ and free-space voxels as $L_f$, which are then added together to compute the total loss $L = L_s + \lambda_f L_f$.
We set $\lambda_f = 1.5$ to achieve better semantic reconstructions. 

%% file: sec/ex_suncg.tex
The artificial data origin of the SUNCG dataset with 38 semantic labels enables full control of the data fusion. 
All components of the method are examined on this dataset with an ablation study. 
We simulate several different depth sensors, such as a perfect sensor, Kinect and different stereo algorithms. 

The baseline is a recent work by Cherabier~\etal~\cite{Cherabier-et-al-ECCV-2018} where they use a simple averaging of input TSDF volumes trained with the network without confidence estimation module. 
We add gradually the following input measurements: average 3$\times$3 depth patches, mean and standard deviation of gradients, mean and standard deviation of normalized cross-correlation between stereo patches in case of a stereo algorithm. 

Training and validation sets were created by randomly selecting 100 and 30 scenes respectively.
Qualitative results for a selected scene and quantitative results on the whole dataset are shown in Fig.~\ref{fig:suncg}. 
Every input brings an increase in performance, measured by semantic and free-space accuracy. 
Quantitative results contain only semantic accuracy. The free-space accuracy was close to 0.95 with small deviations in all settings. 
The increase in accuracy is small, but the values approach the upper limit given by the perfect sensor and the reconstructions look better visually.

%% file: figs/scannet.tex
\newcommand{\scannetHg}{0.2}
\newcommand{\scannetHgg}{0.18}
\begin{figure*} 
\begin{center}
\begin{tabular}{ccc}
\includegraphics[width=\scannetHg\textwidth]{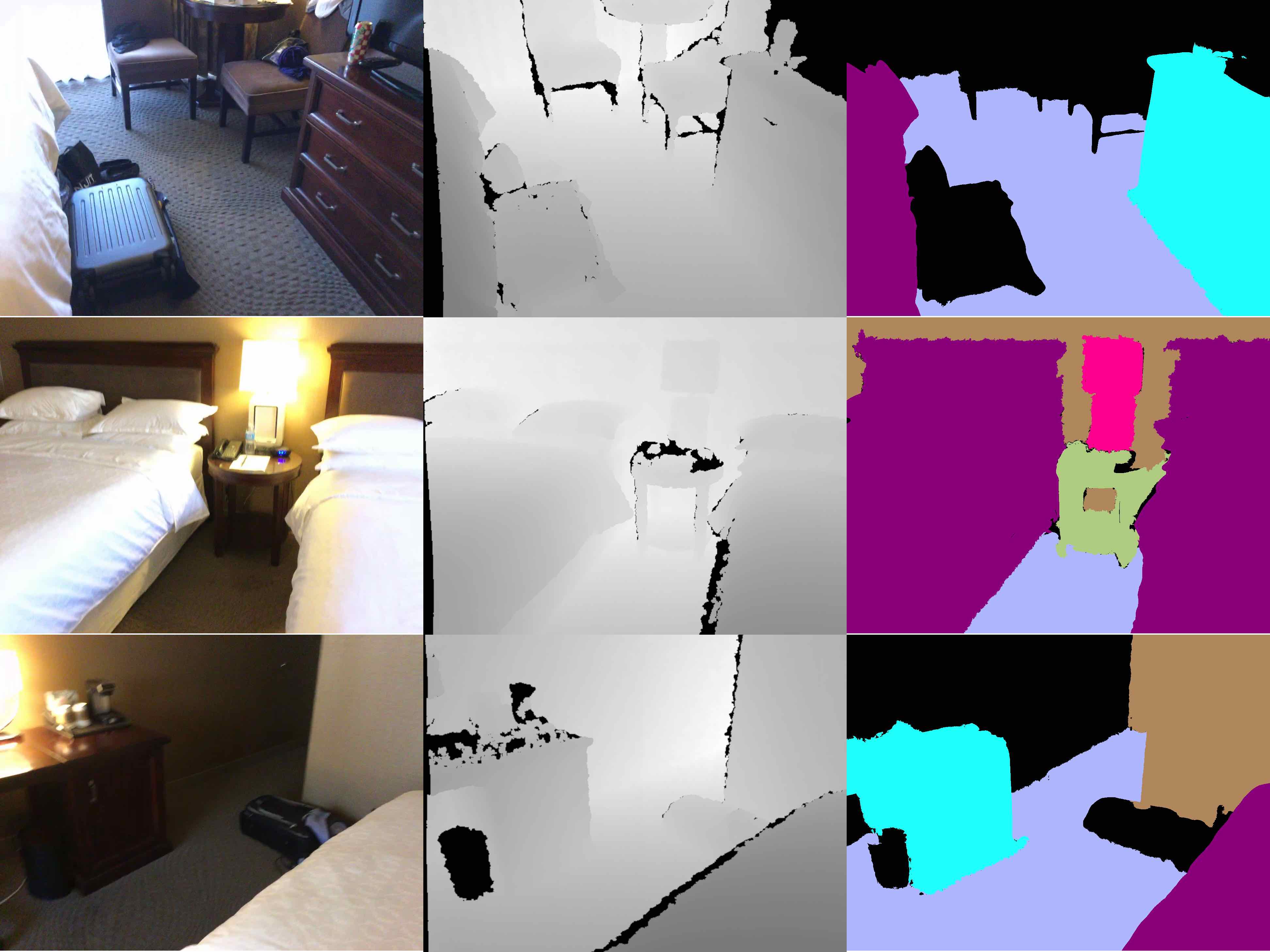} &
\includegraphics[width=\scannetHg\textwidth]{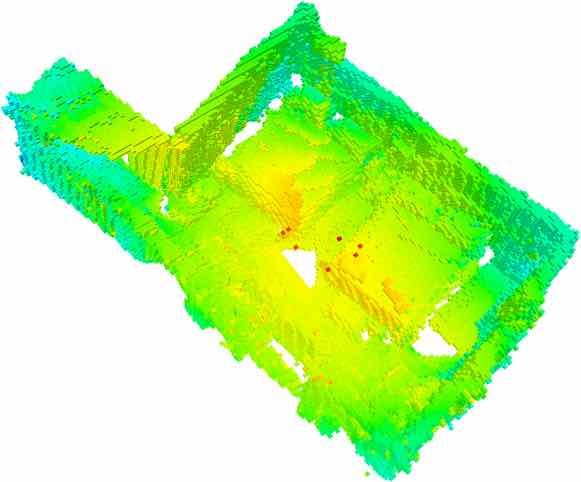} &
\includegraphics[width=\scannetHg\textwidth]{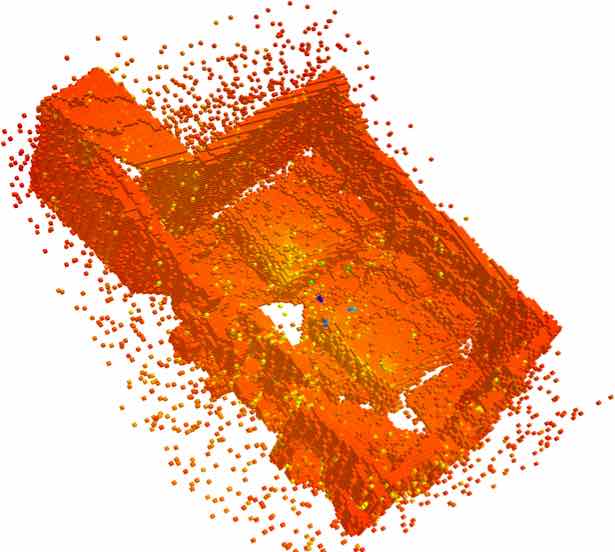} \\
Input Images & Kinect Confidences & Noisy Kinect Confidences \\
\end{tabular}
\begin{tabular}{cccc}
\includegraphics[width=\scannetHg\textwidth]{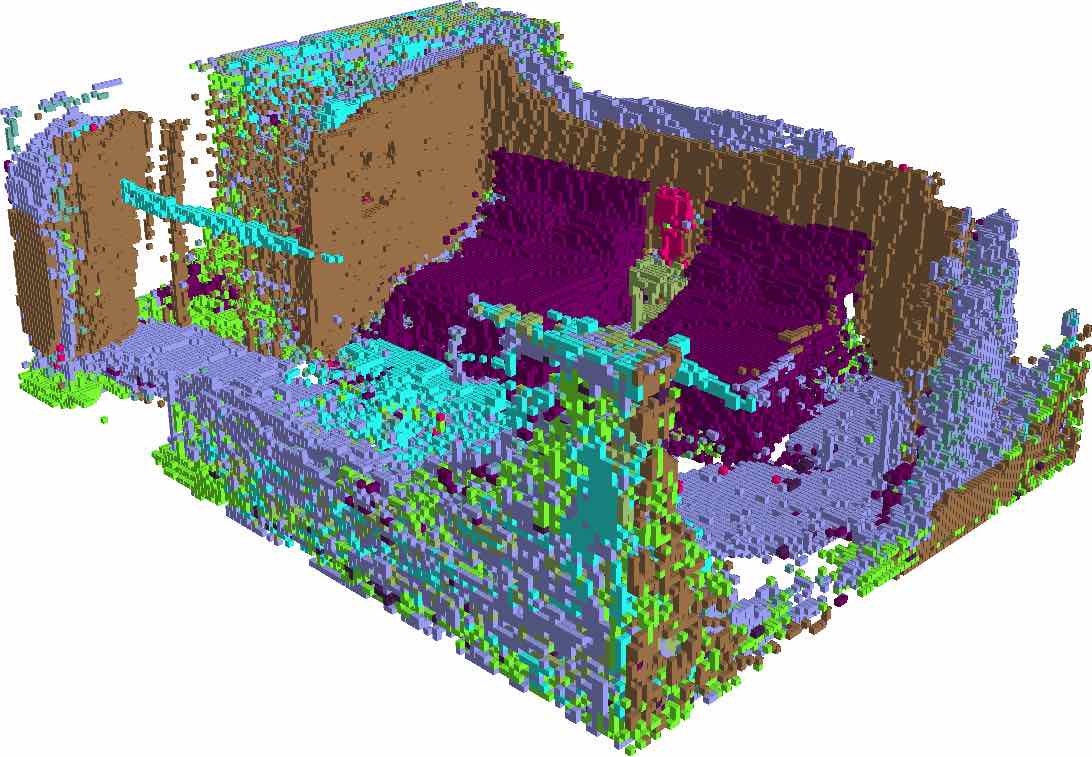} &
\includegraphics[width=\scannetHg\textwidth]{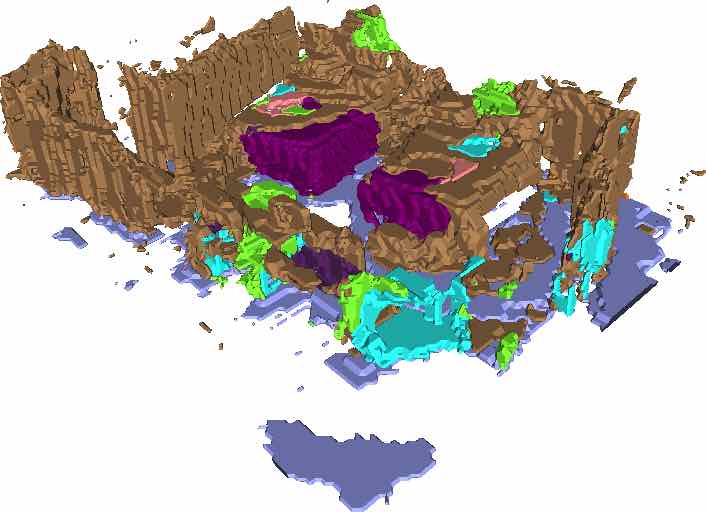} &
\includegraphics[width=\scannetHg\textwidth]{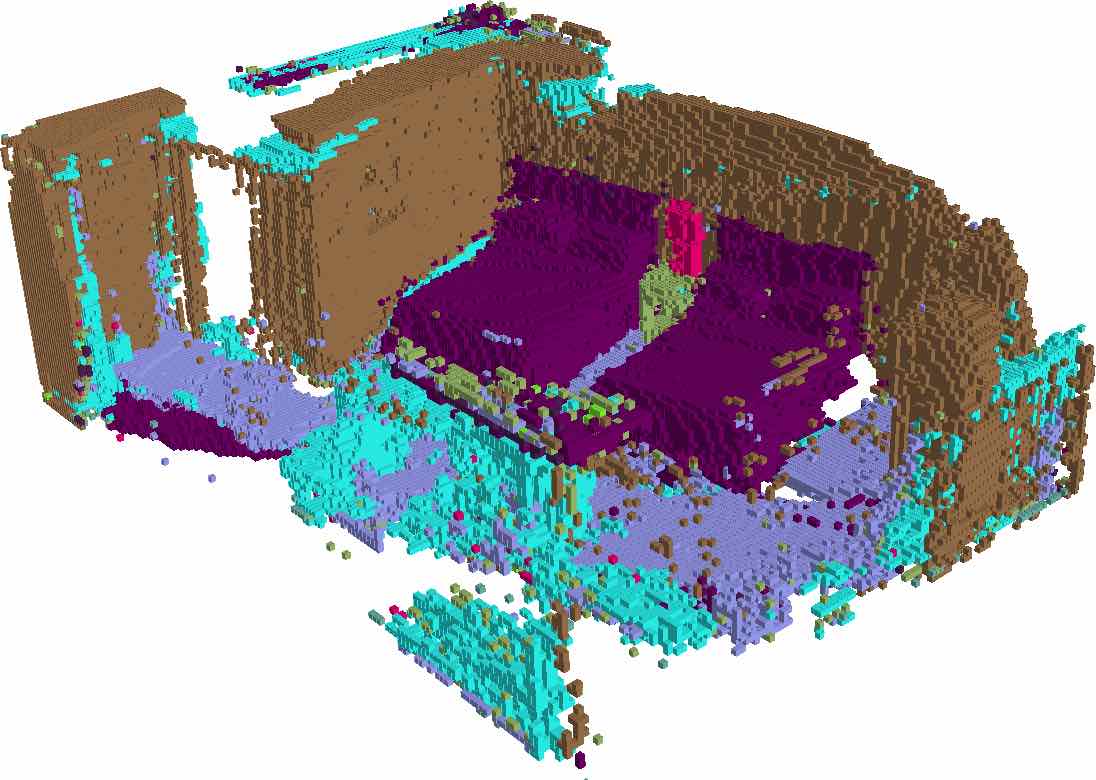} &
\includegraphics[width=\scannetHg\textwidth]{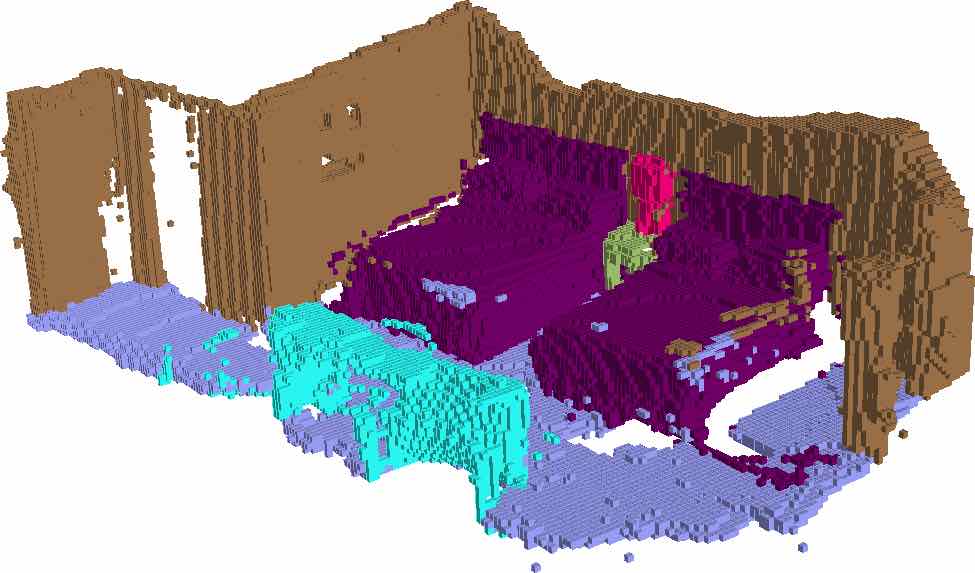} \\
\includegraphics[width=\scannetHgg\textwidth]{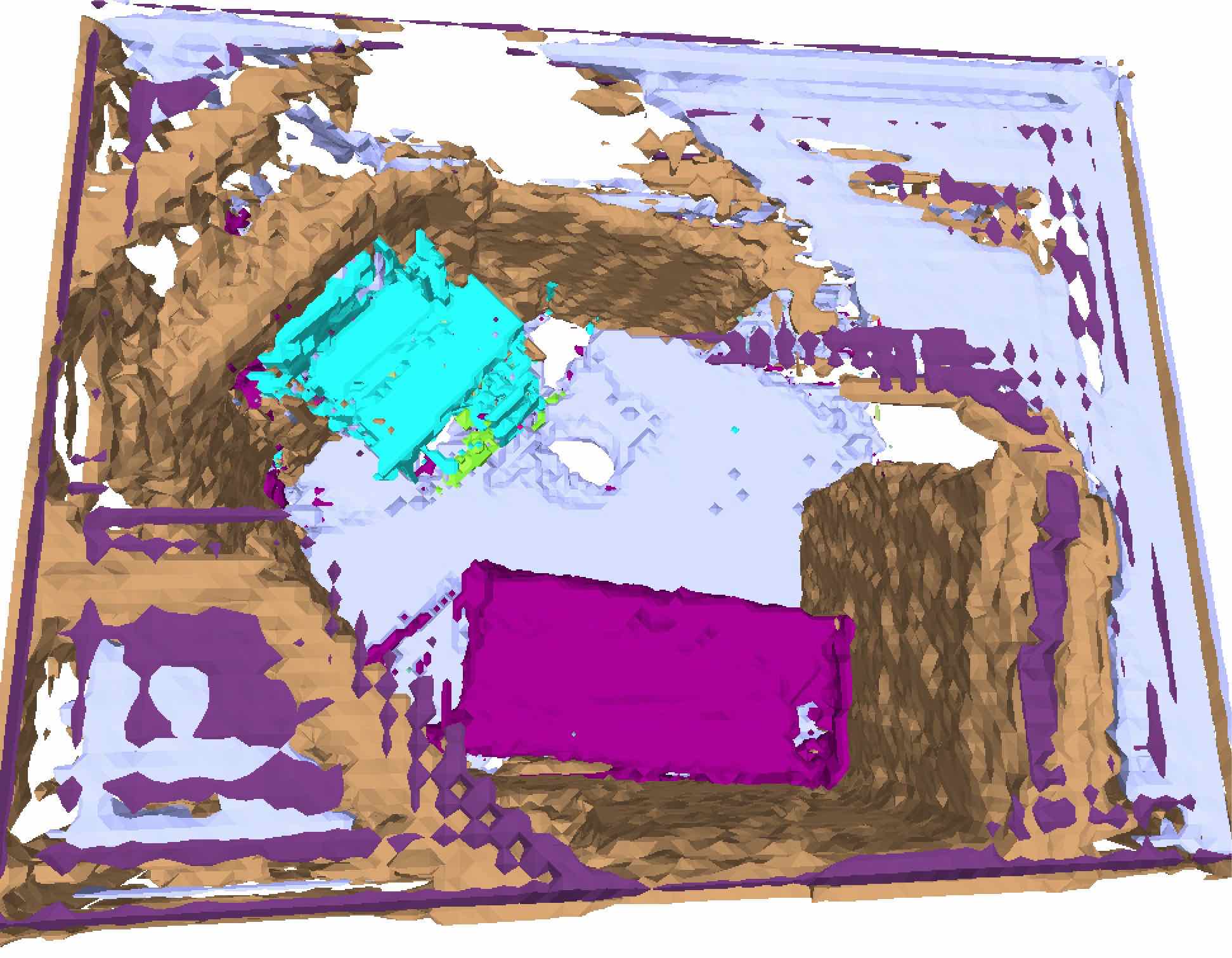} &
\includegraphics[width=\scannetHgg\textwidth]{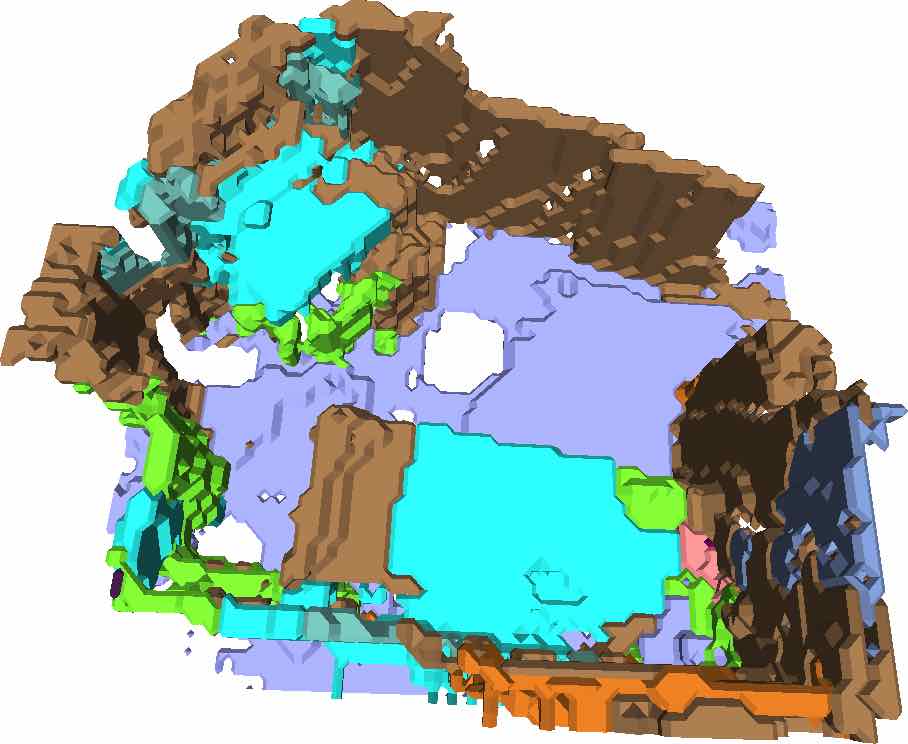} &
\includegraphics[width=\scannetHgg\textwidth]{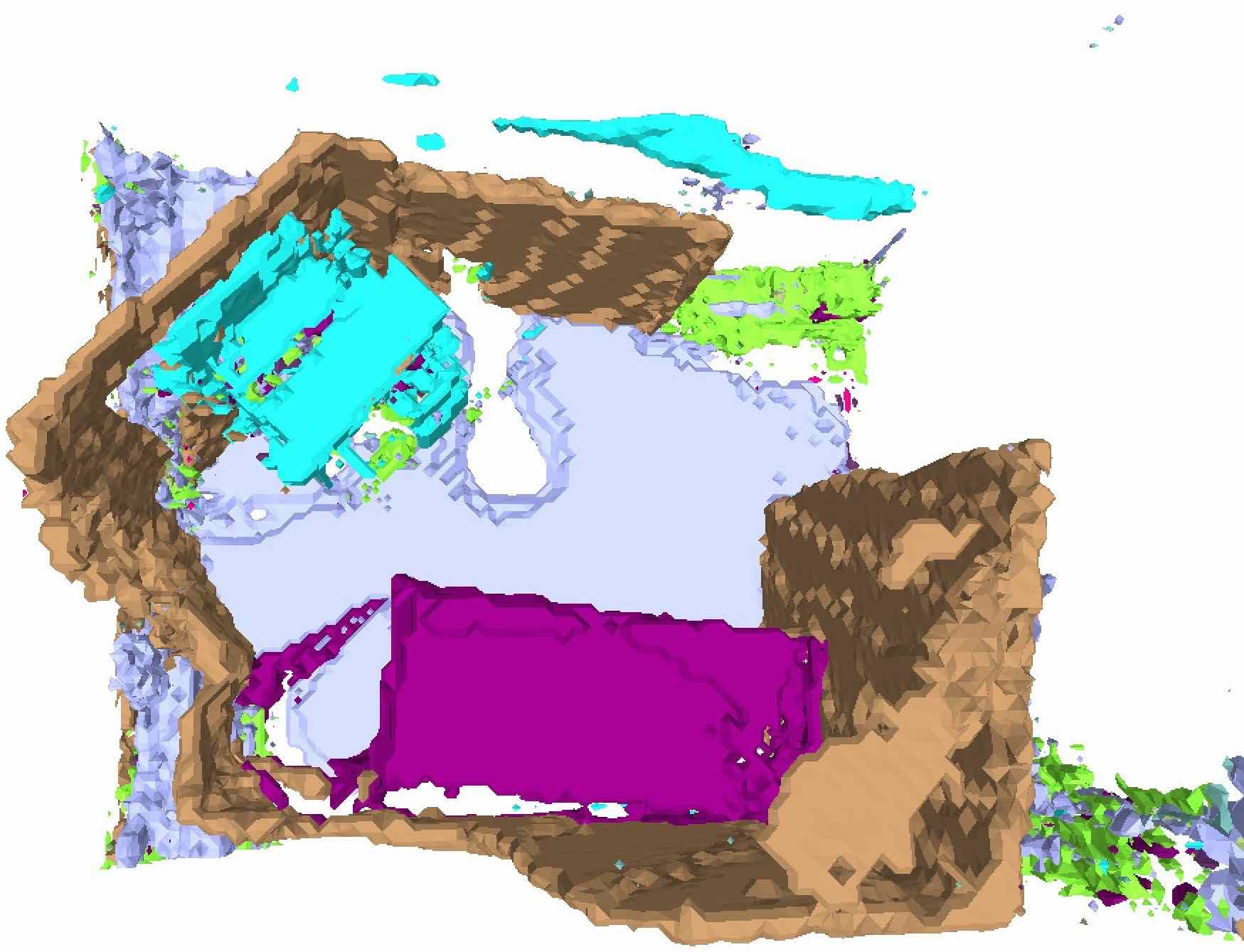} &
\includegraphics[width=\scannetHgg\textwidth]{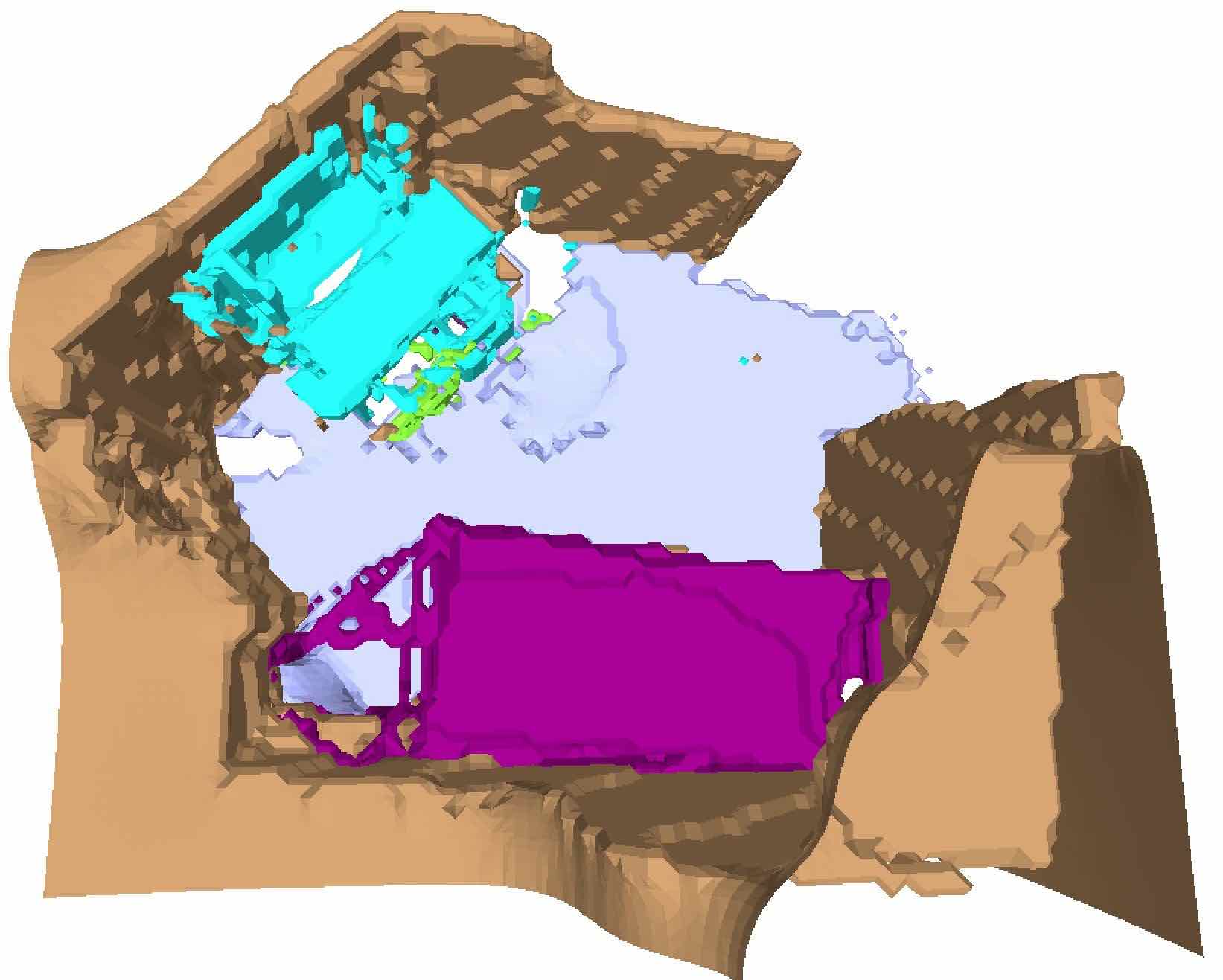} \\
Standard TSDF in \cite{Cherabier-et-al-ECCV-2018} & ScanComplete~\cite{Dai-et-al-CVPR-2018}  & Learned Fusion (ours) & Ground Truth \\
\end{tabular}
\end{center}
\caption{\textbf{Our learned fusion approach in comparison to standard TSDF averaging in~\cite{Cherabier-et-al-ECCV-2018}.} We fuse real world Kinect data with an artificially noised Kinect sensor. Our approach leads to less artificats, more consistent semantic labels and better accuracy scores in comparison to the ground truth data. \textbf{Top row}: Estimated confidence values from input measurement on ScanNet dataset. \textbf{Middle and bottom rows}: The proposed learned fusion in comparison to datacost averaging in~\cite{Cherabier-et-al-ECCV-2018}, ScanComplete~\cite{Dai-et-al-CVPR-2018} and ground truth.}\label{fig:scannet}
\end{figure*}

%% file: sec/ex_expert_system.tex
The proposed method was applied to create an expert system for stereo algorithms. 
We used the following four methods for stereo depth estimation: 
\begin{itemize}[topsep=2pt,leftmargin=*]
\setlength\itemsep{0mm}
\item Pyramid Stereo Matching Network (PSMNet)~\cite{psmnet}~--~ 3D CNN architecture with spatial pyramid pooling module for depth map estimation from a stereo pair.
\item Depth Prediction with Fully Convolutional Residual Networks (FCRN)~\cite{fcrn}~--~fully convolutional architecture with residual learning which is trained to estimate depth map from a single RGB image.
\item Semi-Global Block Matching (SGBM)~\cite{sgbm}~--~classical method (by H. Hirschmuller), which matches blocks of a given size in a pair of images using mutual information. 
\item Block Matching (BM)~\cite{opencv_library}~--~a version of block matching algorithm provided by K. Konolige.
\end{itemize}
At first, we trained a network without confidence values on each of the stereo algorithms separately.
Then, a fused combination of these methods with learned confidence values was trained. 
Fig.~\ref{fig:expert} shows that the learned fusion performs better than any of the stereo methods on its own. 
More importantly, the learned fusion results are less noisy, more accurate and complete. 
The stereo system results can be compared to results of other sensor fusion models in Fig.~\ref{fig:suncg}. 

%% file: figs/eth3d_res.tex
\newcommand{\ethHg}{0.10}
\newcommand{\ethWg}{0.13}
\begin{figure*} 
\centering
\begin{tabular}{ccccc}
\includegraphics[height=\ethHg\textwidth]{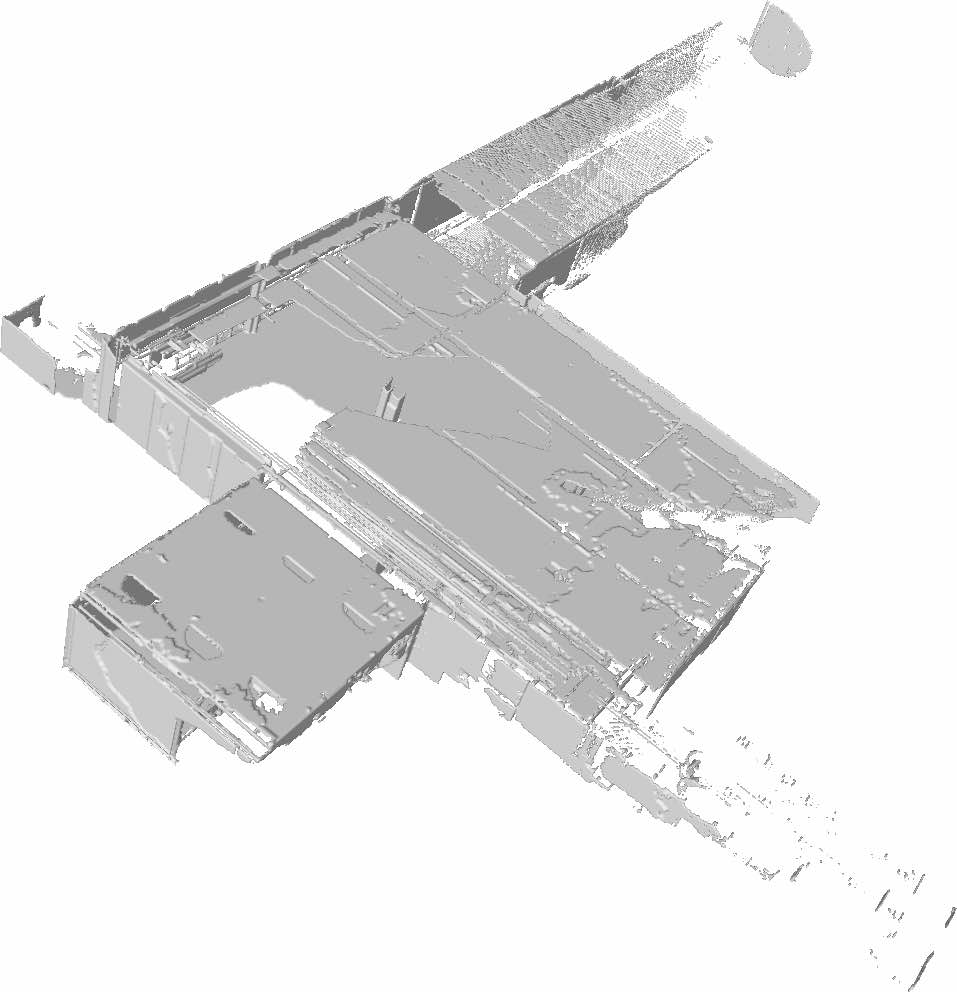} &
\includegraphics[height=\ethHg\textwidth]{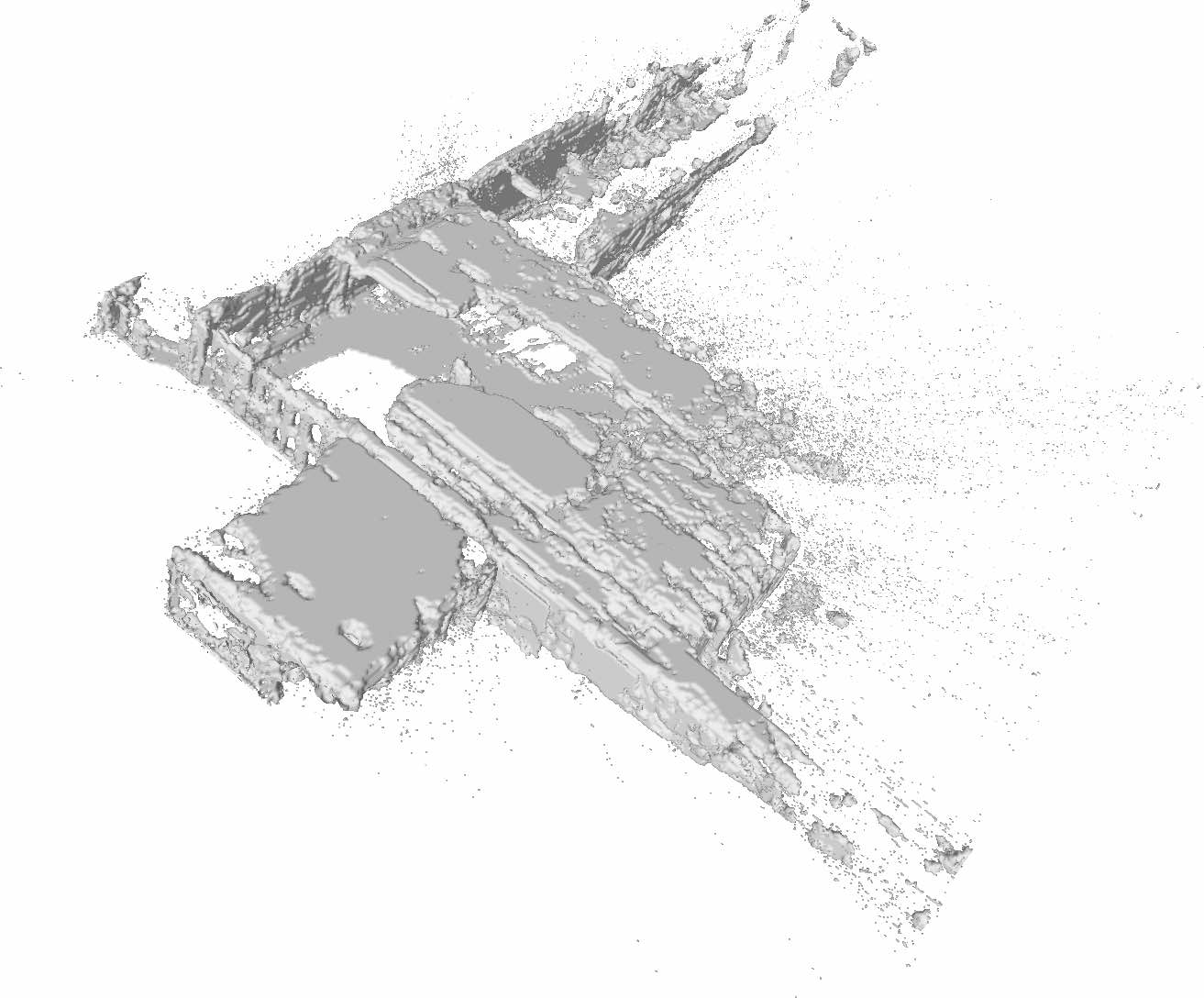} &
\includegraphics[height=\ethHg\textwidth]{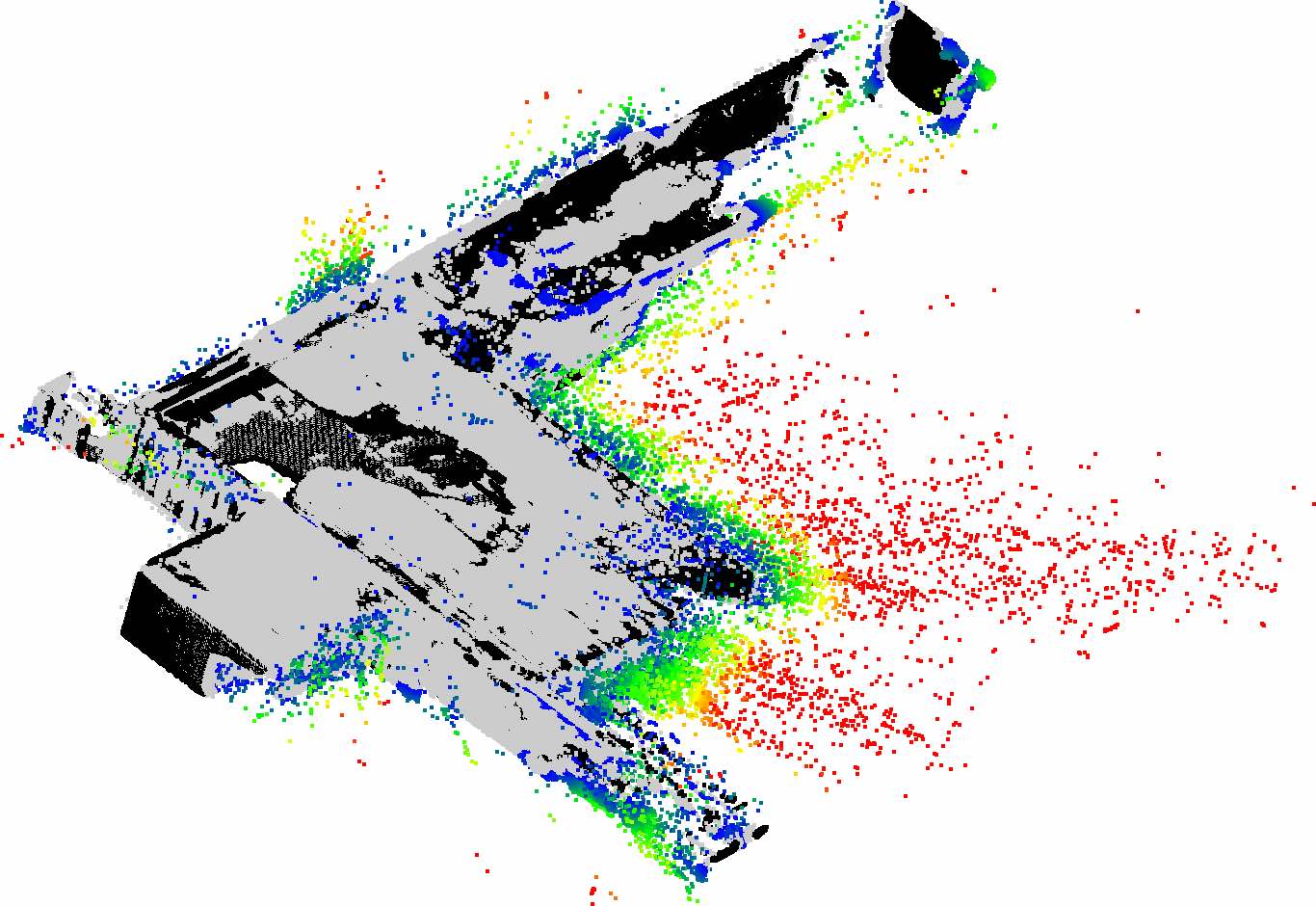} &
\includegraphics[height=\ethHg\textwidth]{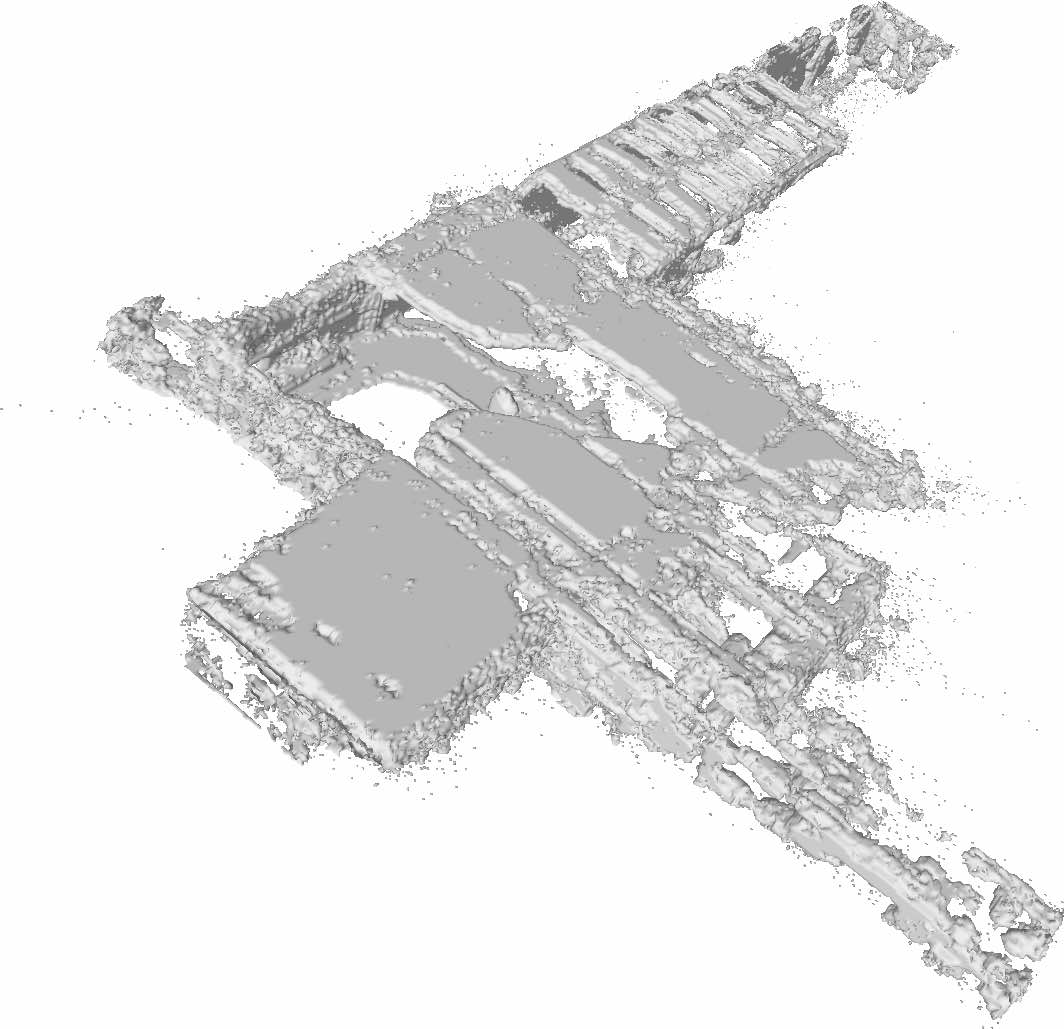} &
\includegraphics[height=\ethHg\textwidth]{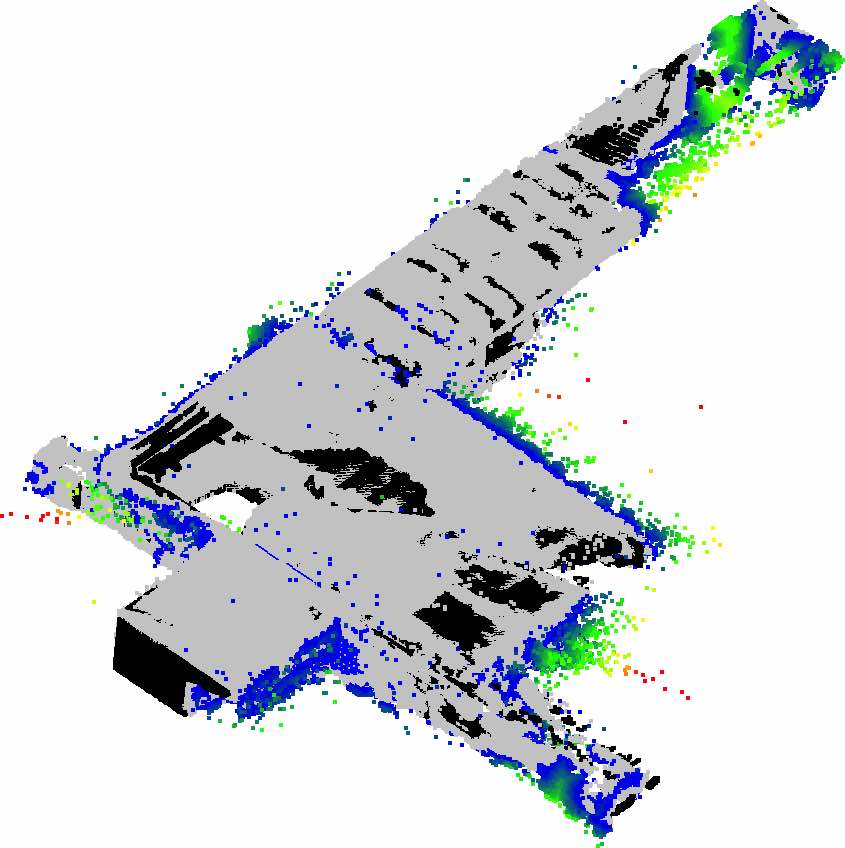} \\

\includegraphics[width=\ethWg\textwidth]{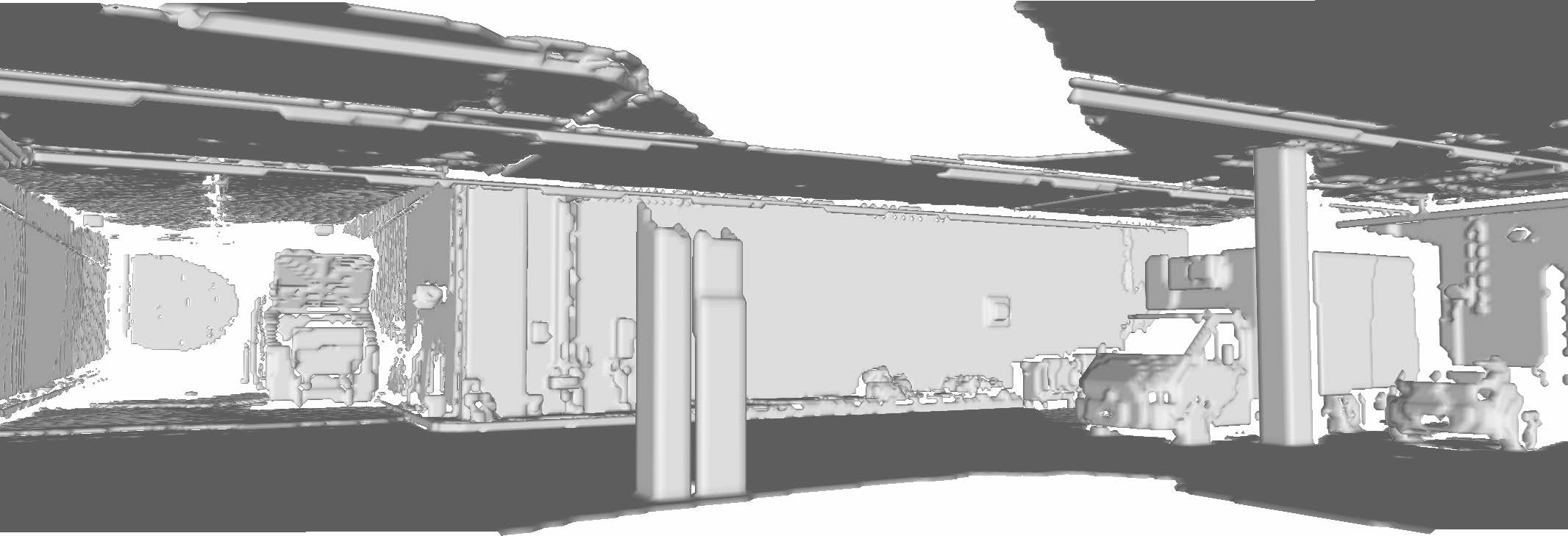} &
\includegraphics[width=\ethWg\textwidth]{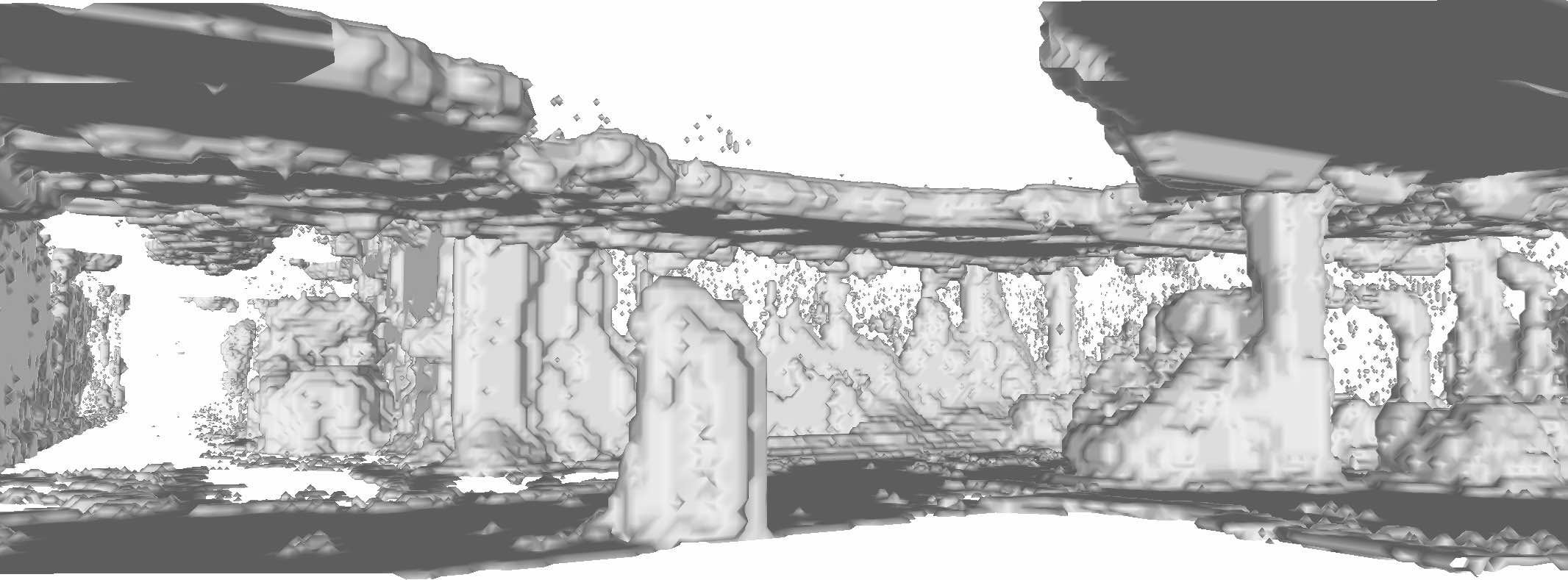} &
\includegraphics[width=\ethWg\textwidth]{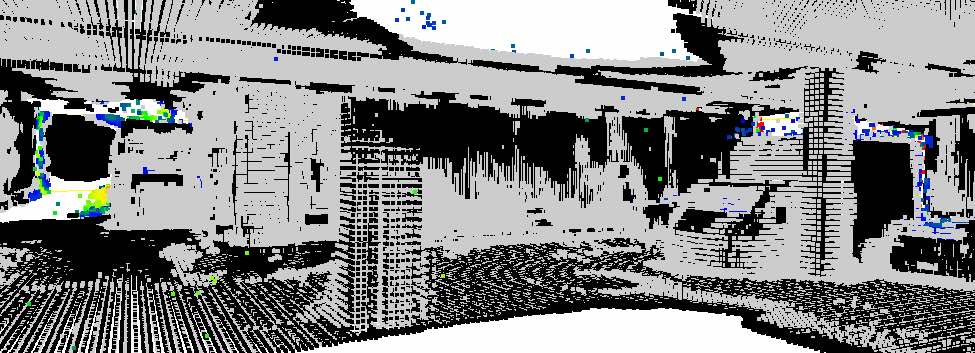} &
\includegraphics[width=\ethWg\textwidth]{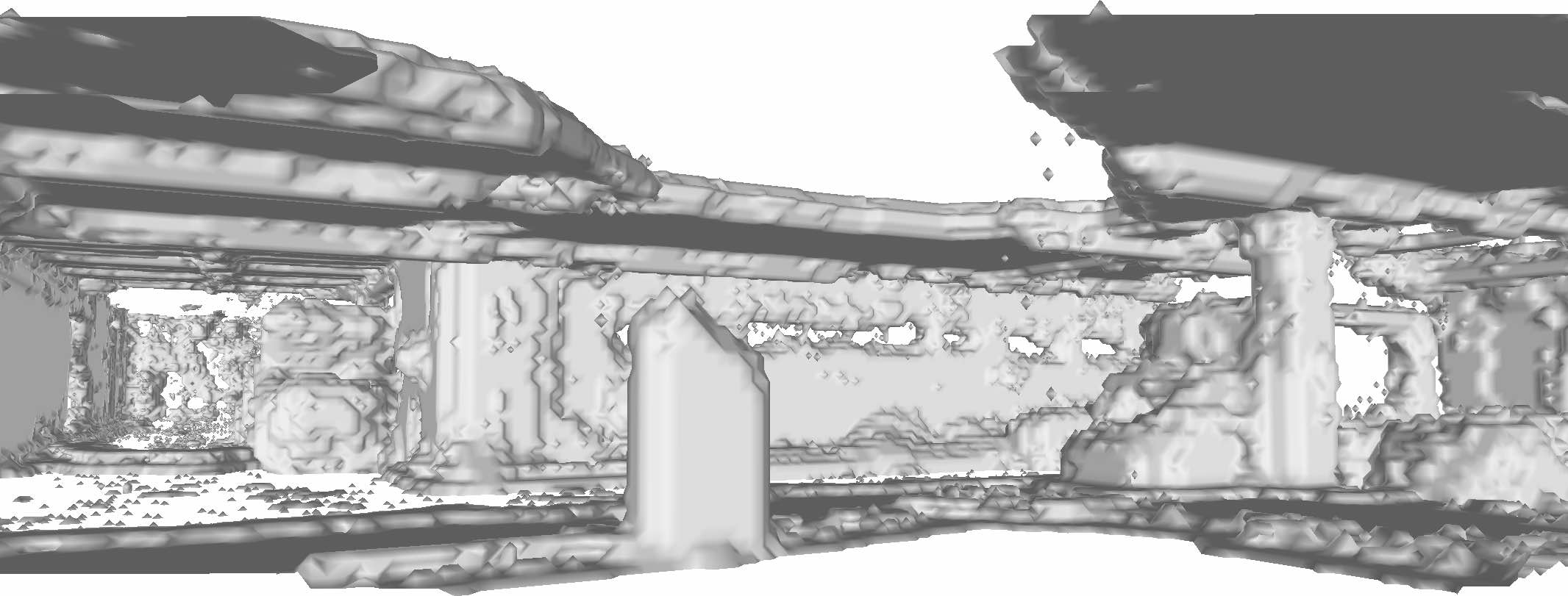} &
\includegraphics[width=\ethWg\textwidth]{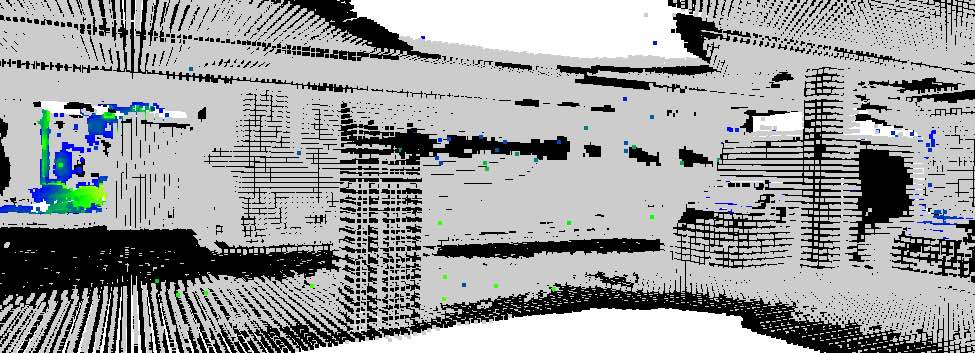} \\

\includegraphics[height=\ethHg\textwidth]{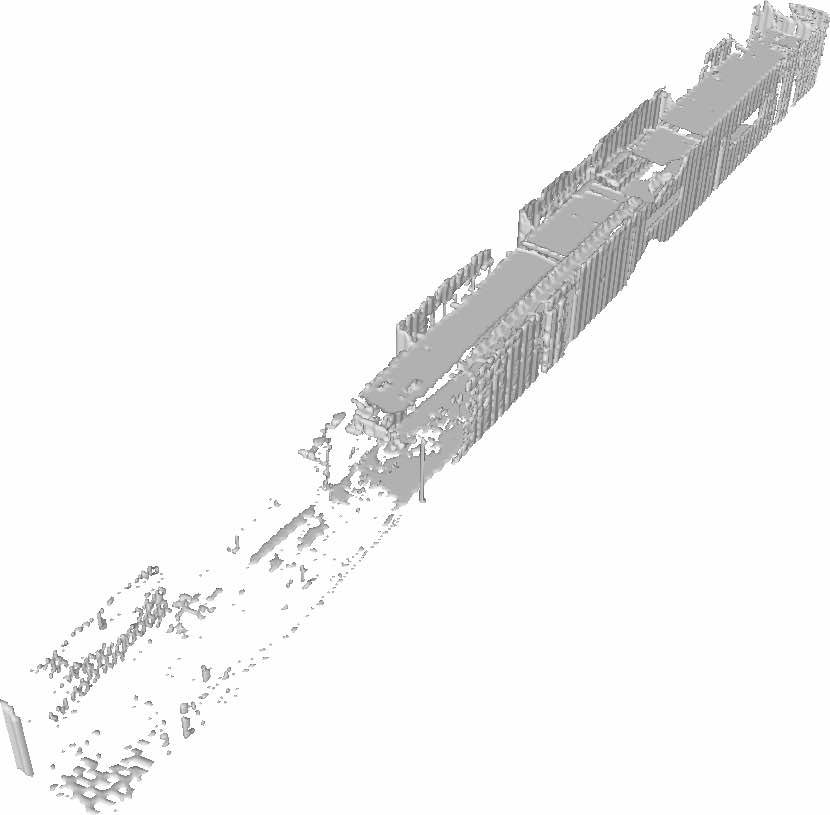} &
\includegraphics[height=\ethHg\textwidth]{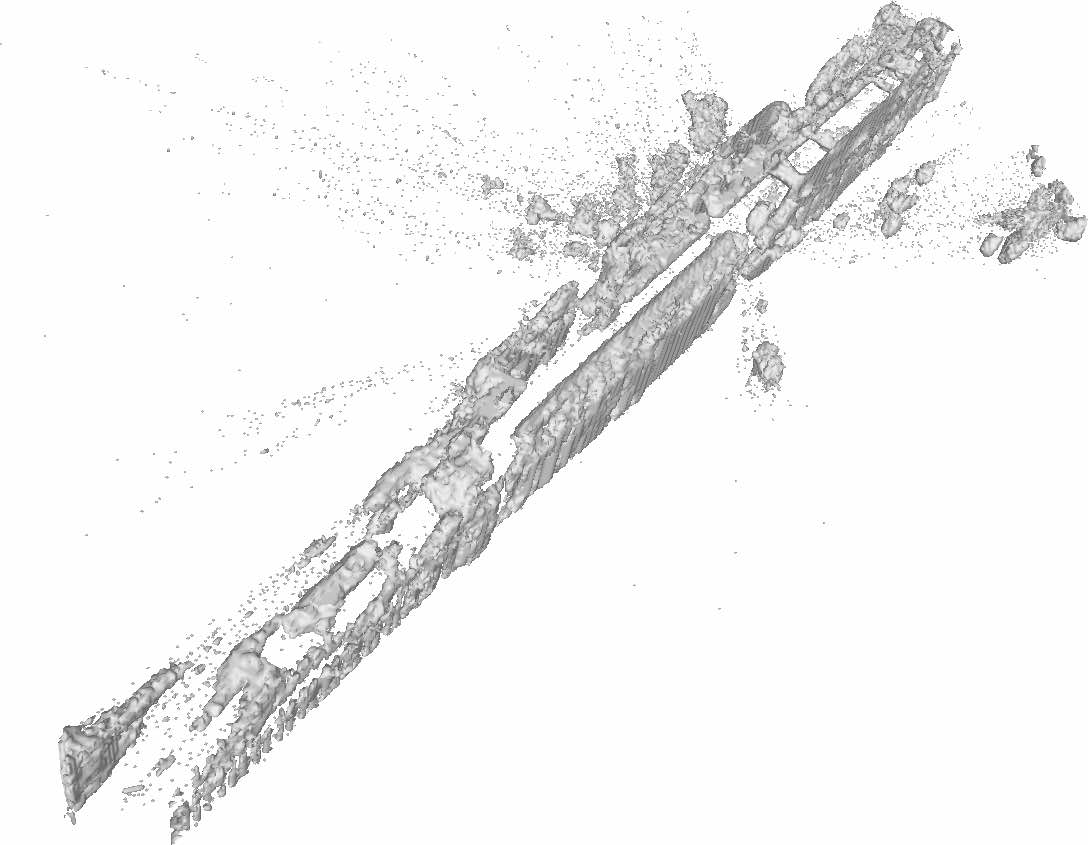} &
\includegraphics[height=\ethHg\textwidth]{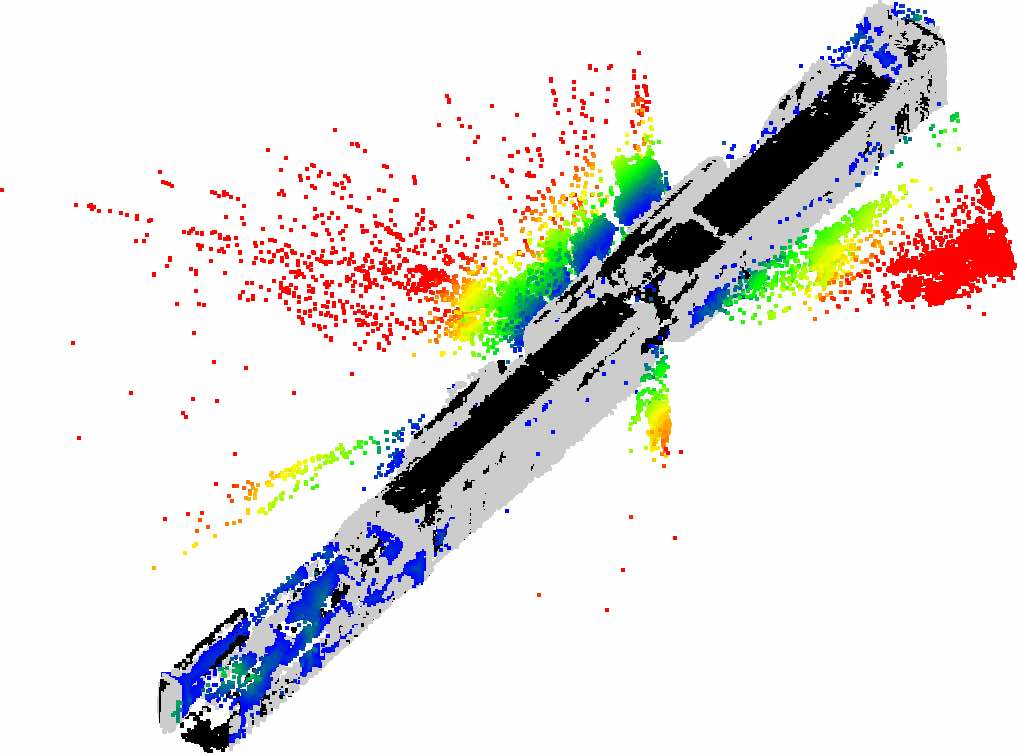} &
\includegraphics[height=\ethHg\textwidth]{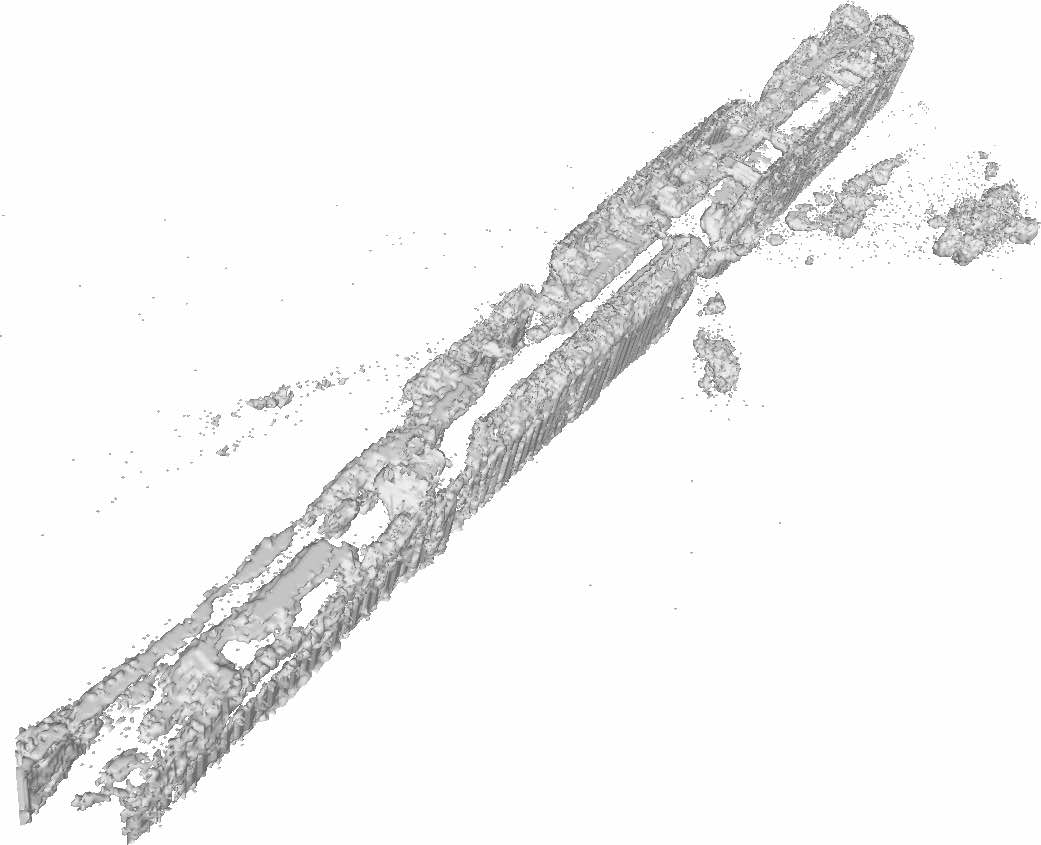} &
\includegraphics[height=\ethHg\textwidth]{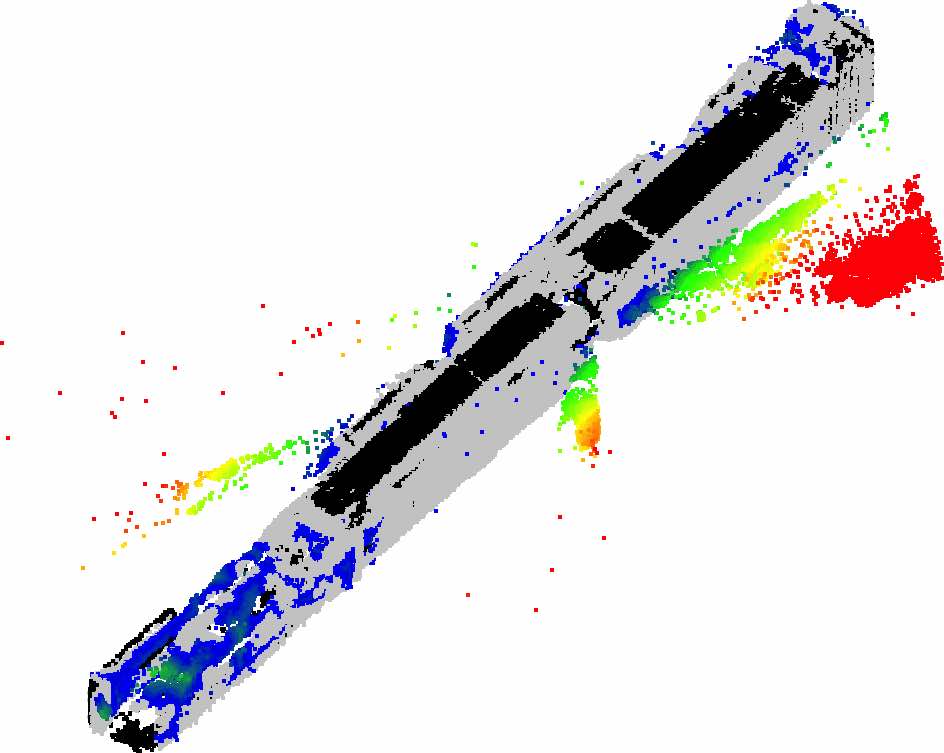} \\

\includegraphics[width=\ethWg\textwidth]{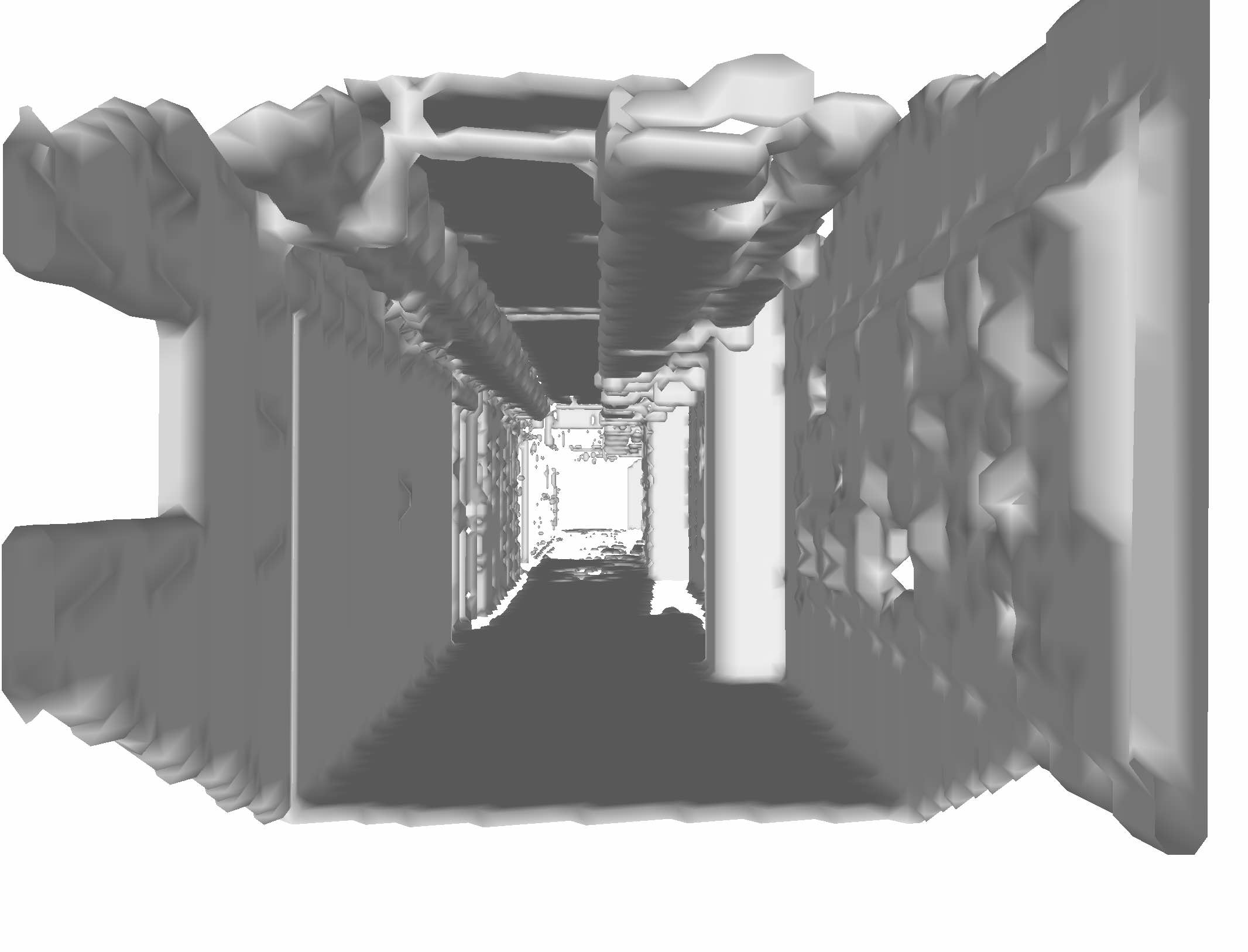} &
\includegraphics[width=\ethWg\textwidth]{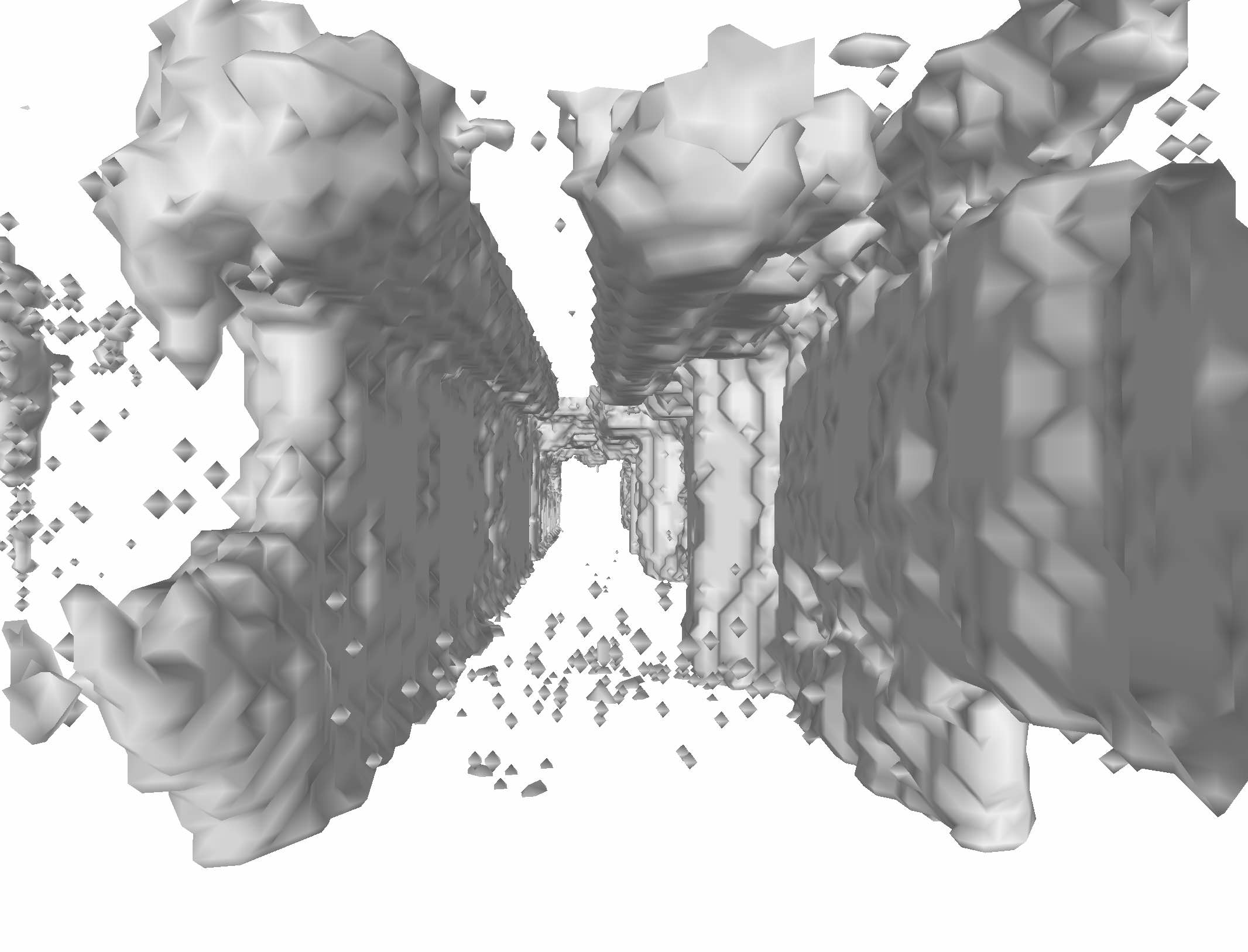} &
\includegraphics[width=\ethWg\textwidth]{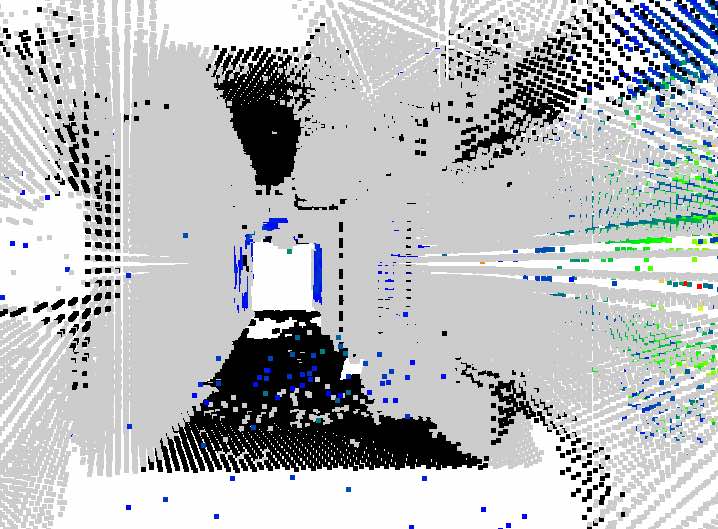} &
\includegraphics[width=\ethWg\textwidth]{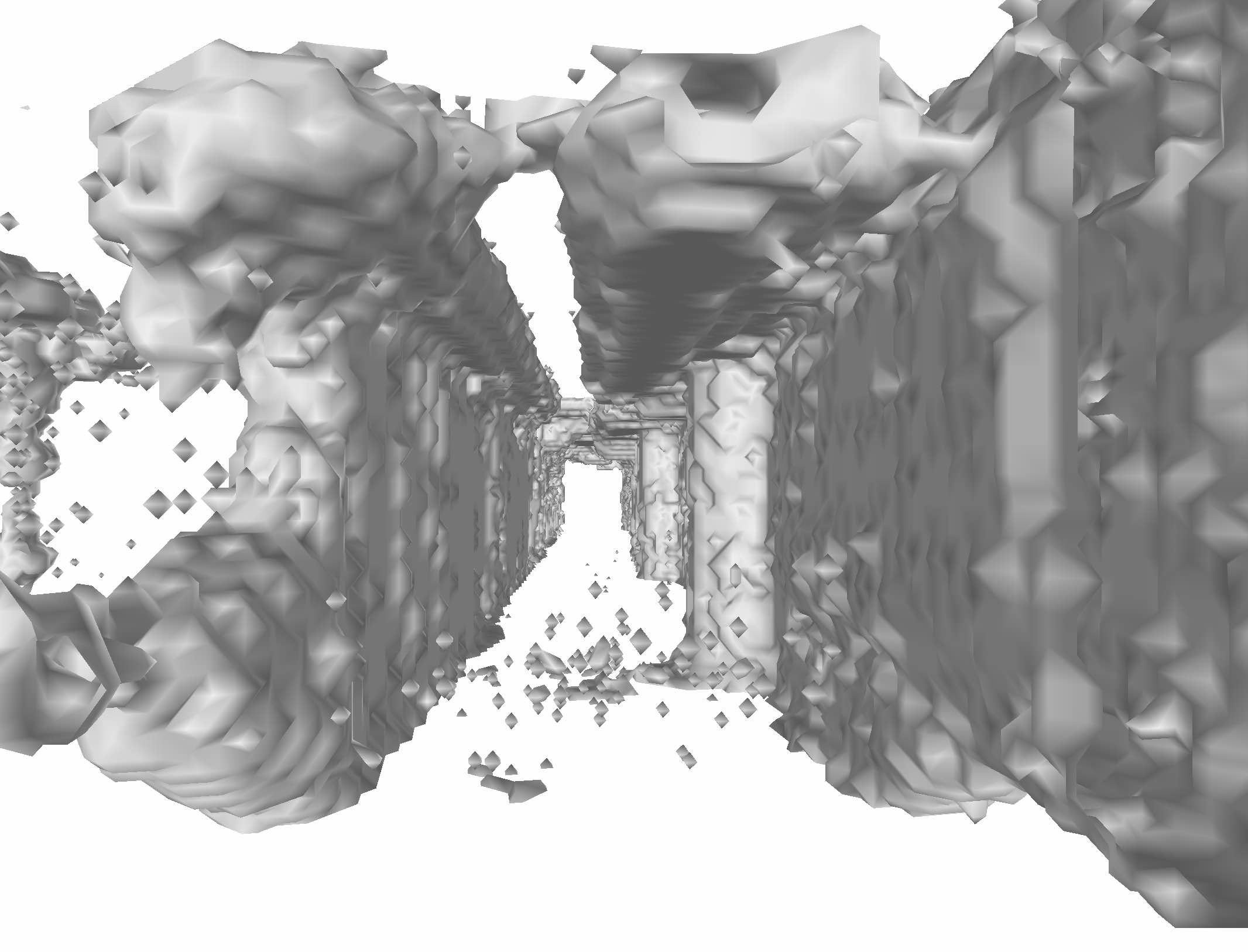} &
\includegraphics[width=\ethWg\textwidth]{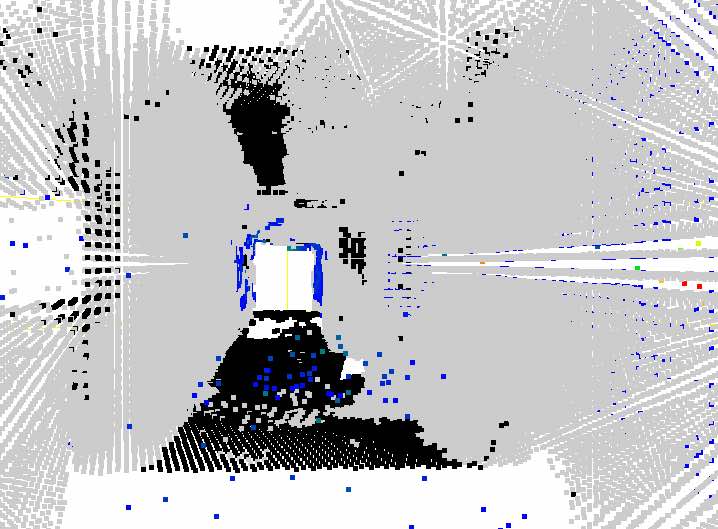} \\

\includegraphics[height=\ethHg\textwidth]{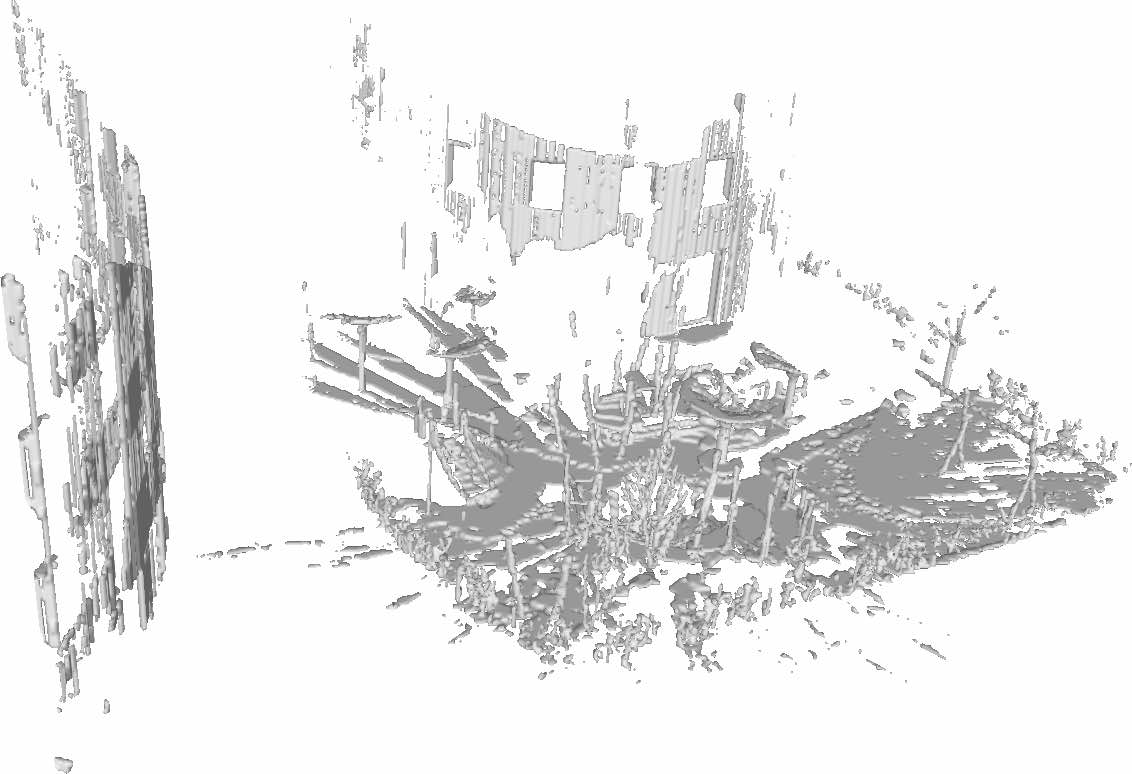} &
\includegraphics[height=\ethHg\textwidth]{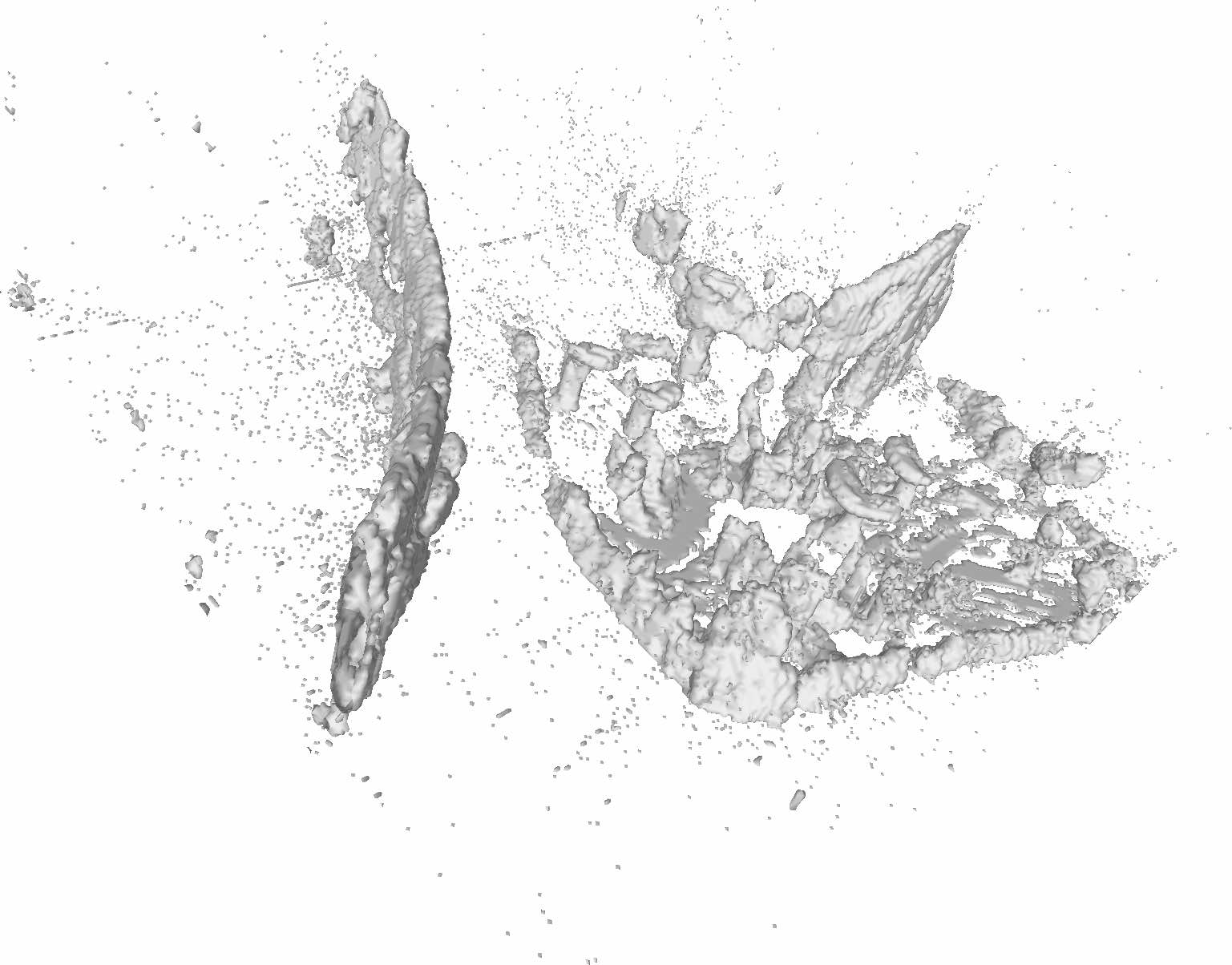} &
\includegraphics[height=\ethHg\textwidth]{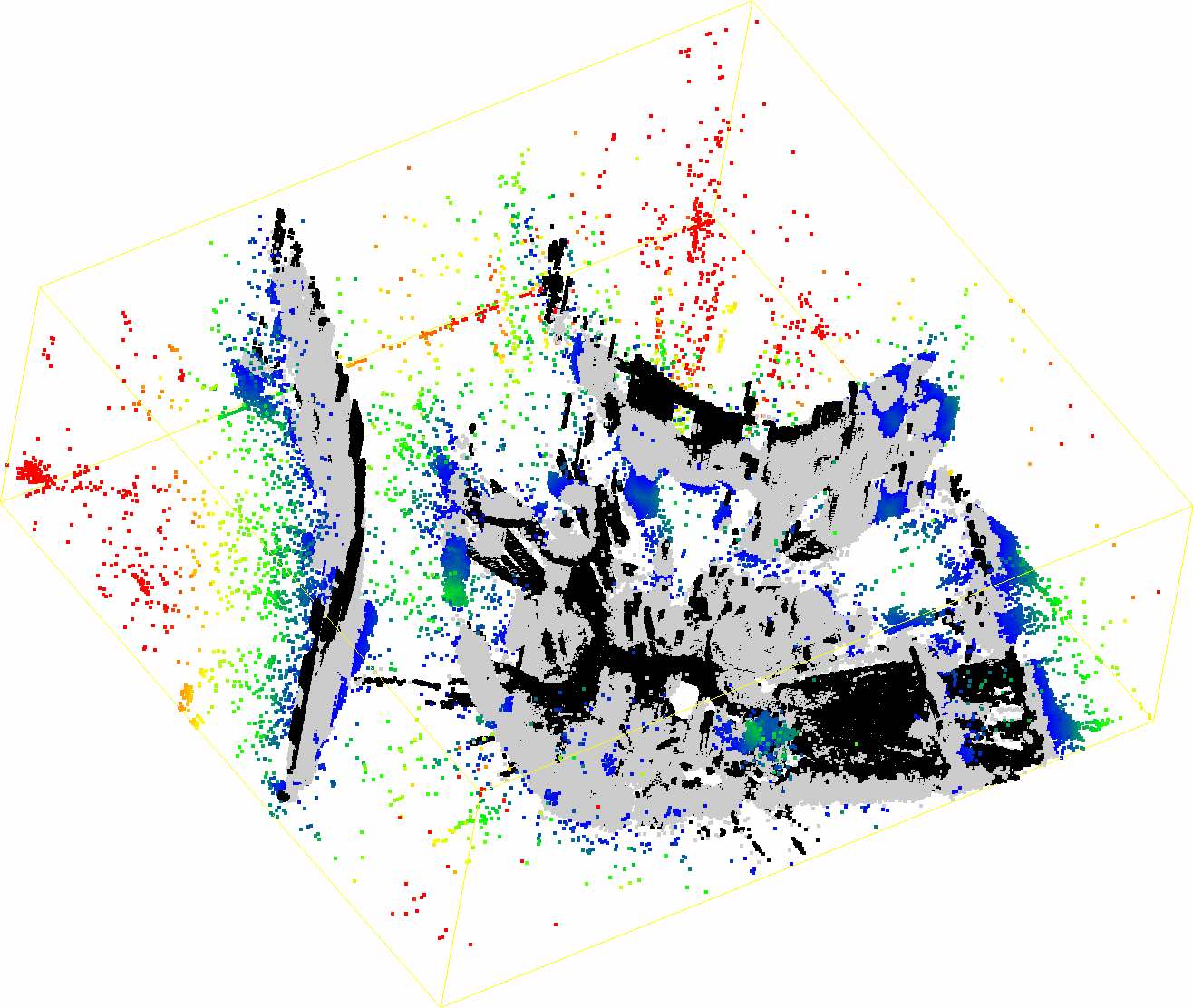} &
\includegraphics[height=\ethHg\textwidth]{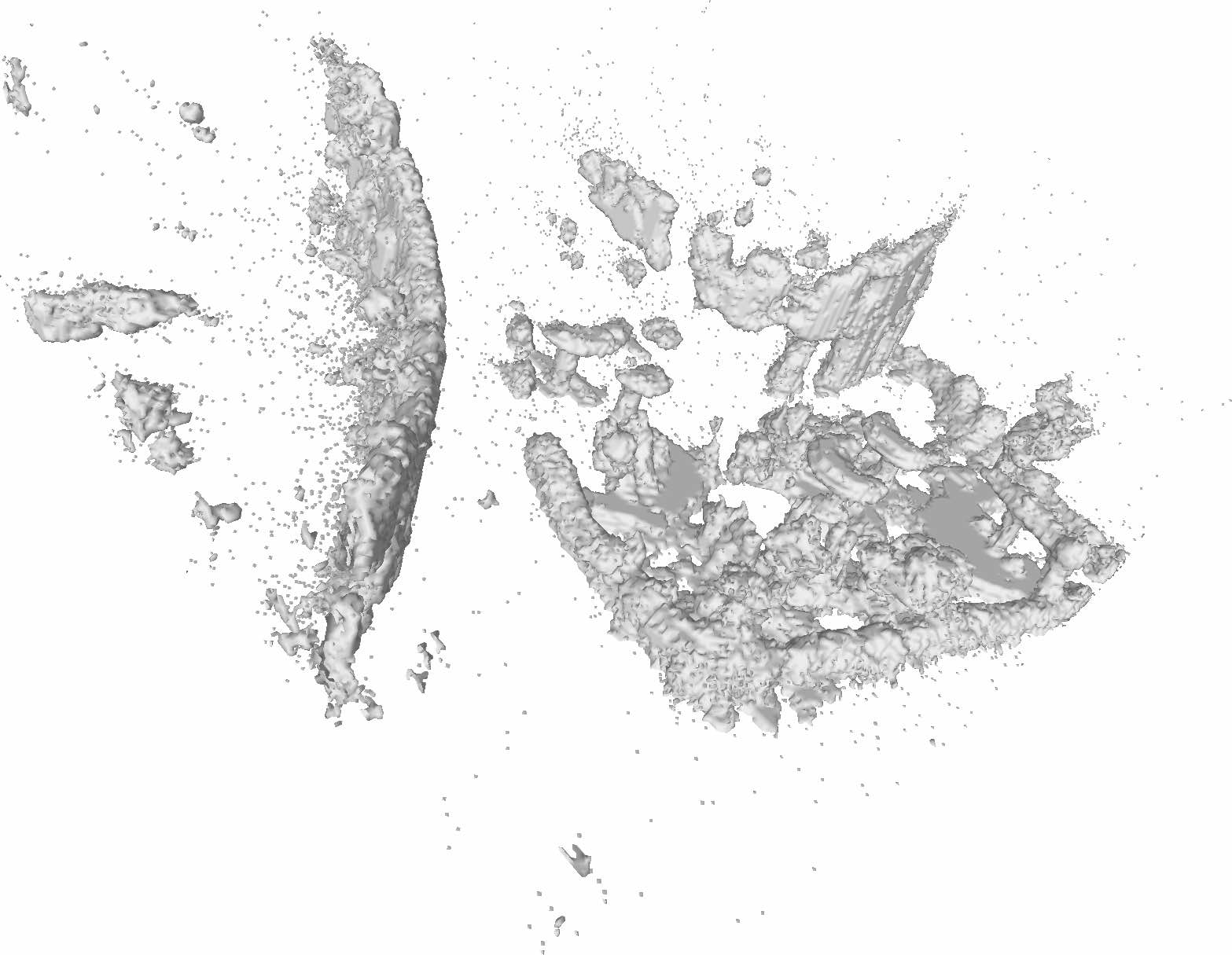} &
\includegraphics[height=\ethHg\textwidth]{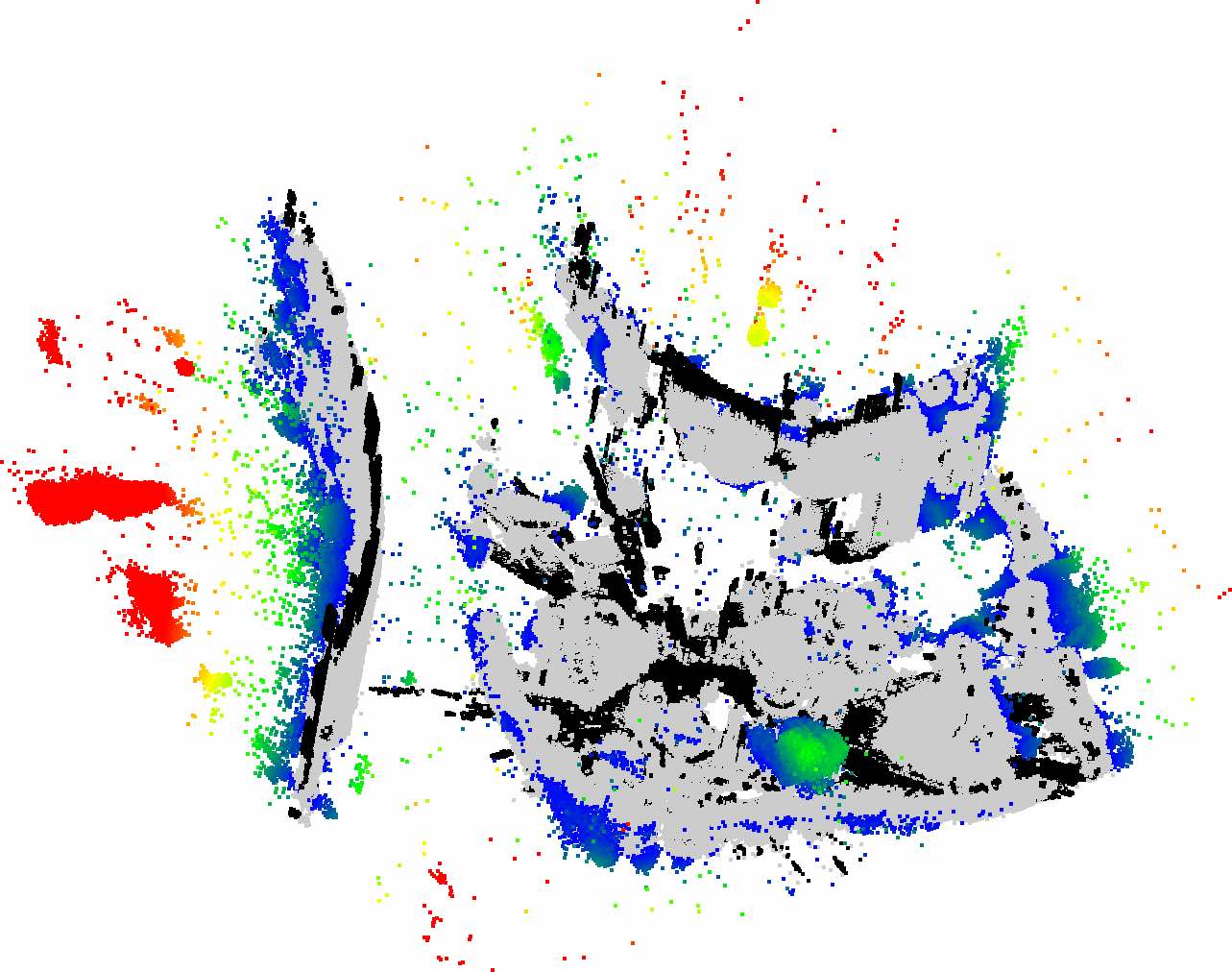} \\

\includegraphics[width=\ethWg\textwidth]{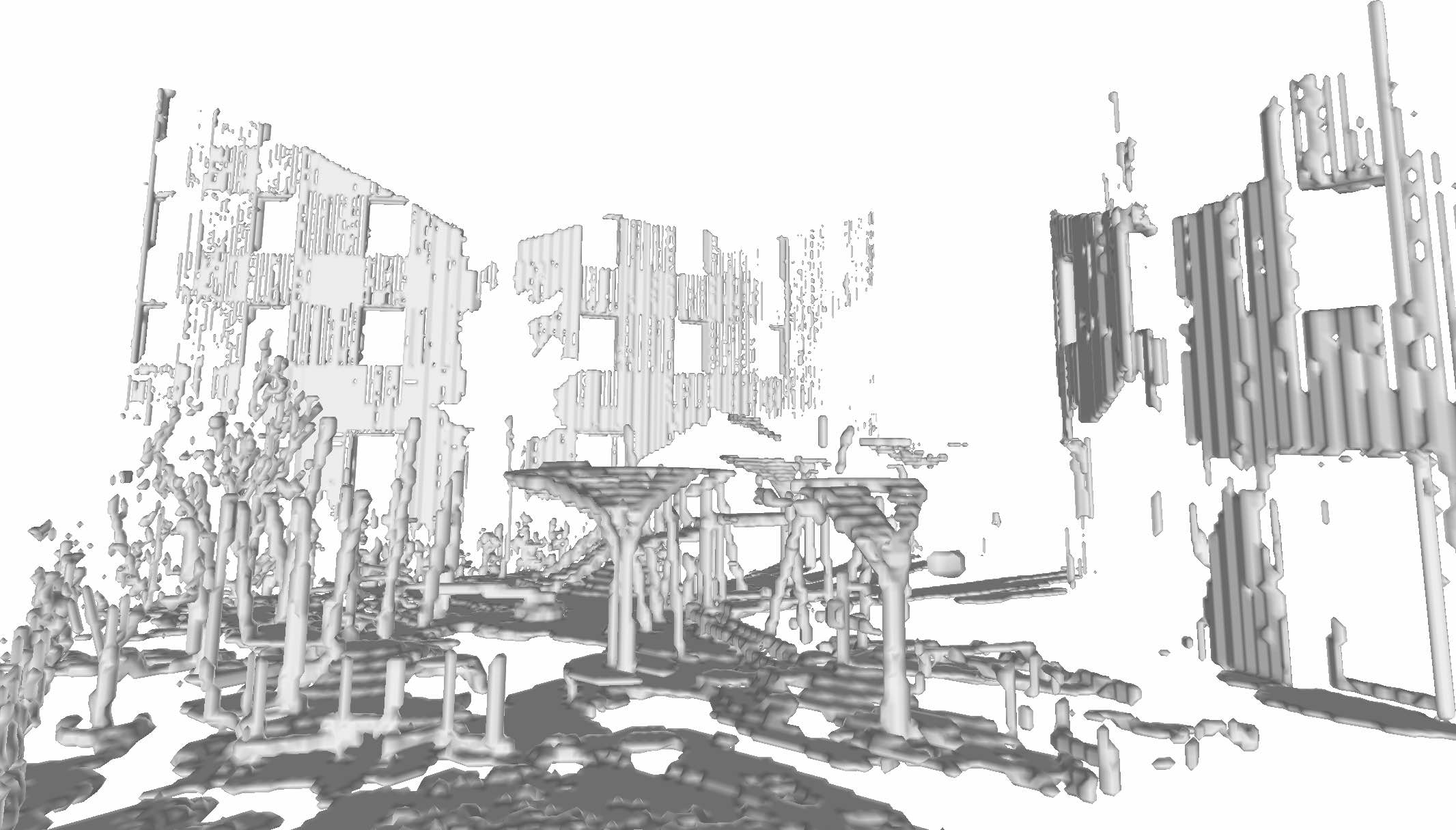} & 
\includegraphics[width=\ethWg\textwidth]{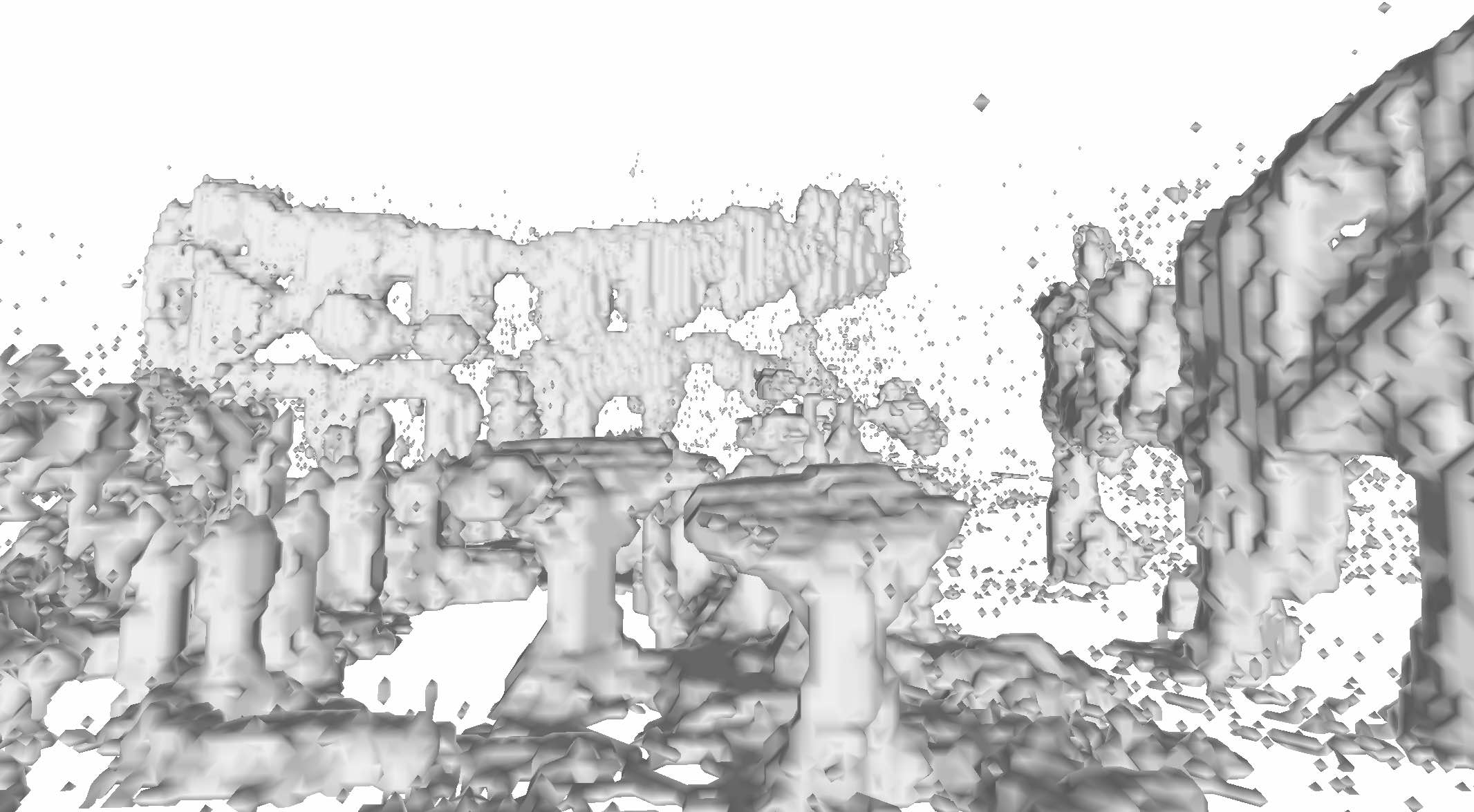} & 
\includegraphics[width=\ethWg\textwidth]{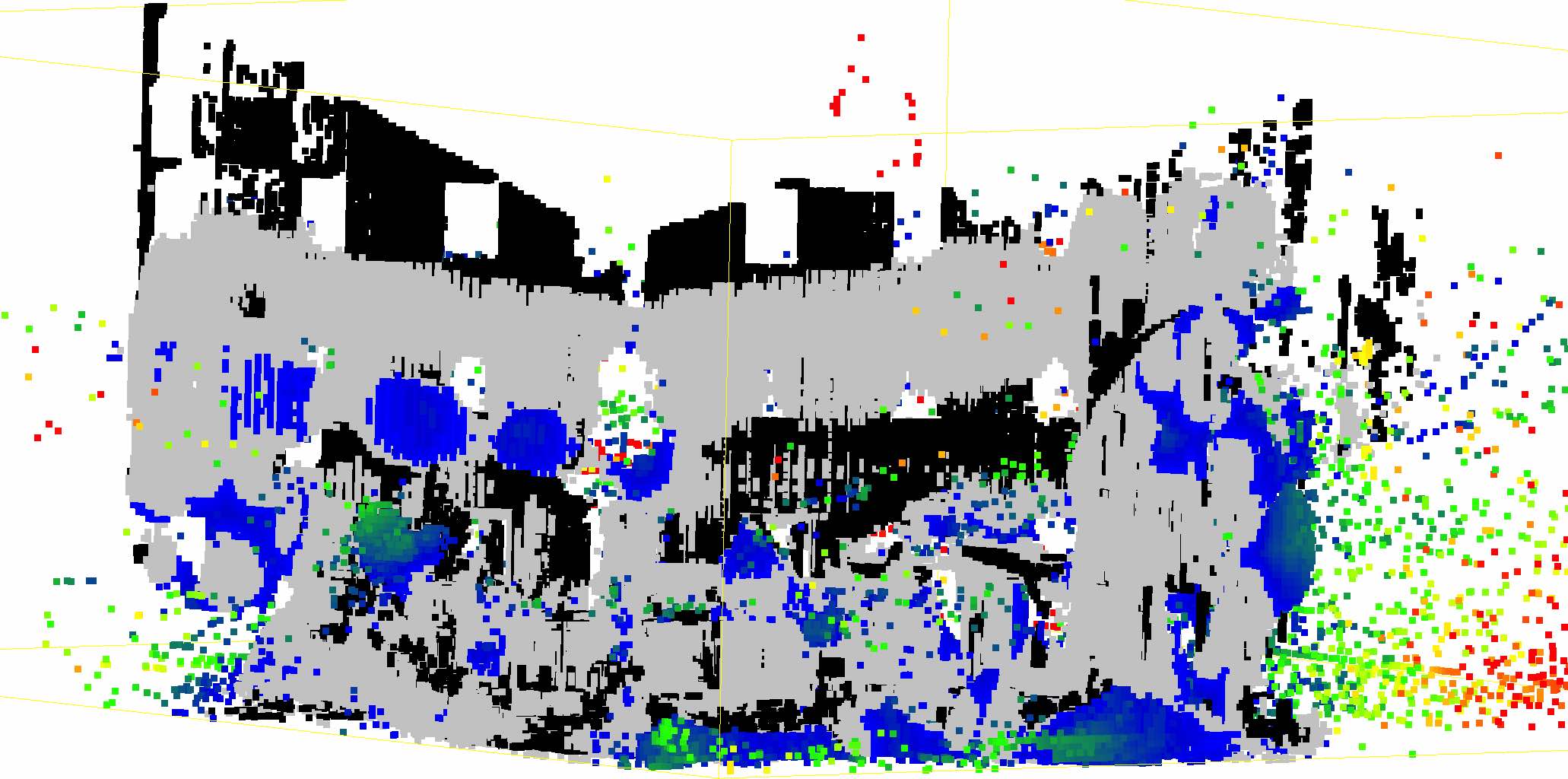} & 
\includegraphics[width=\ethWg\textwidth]{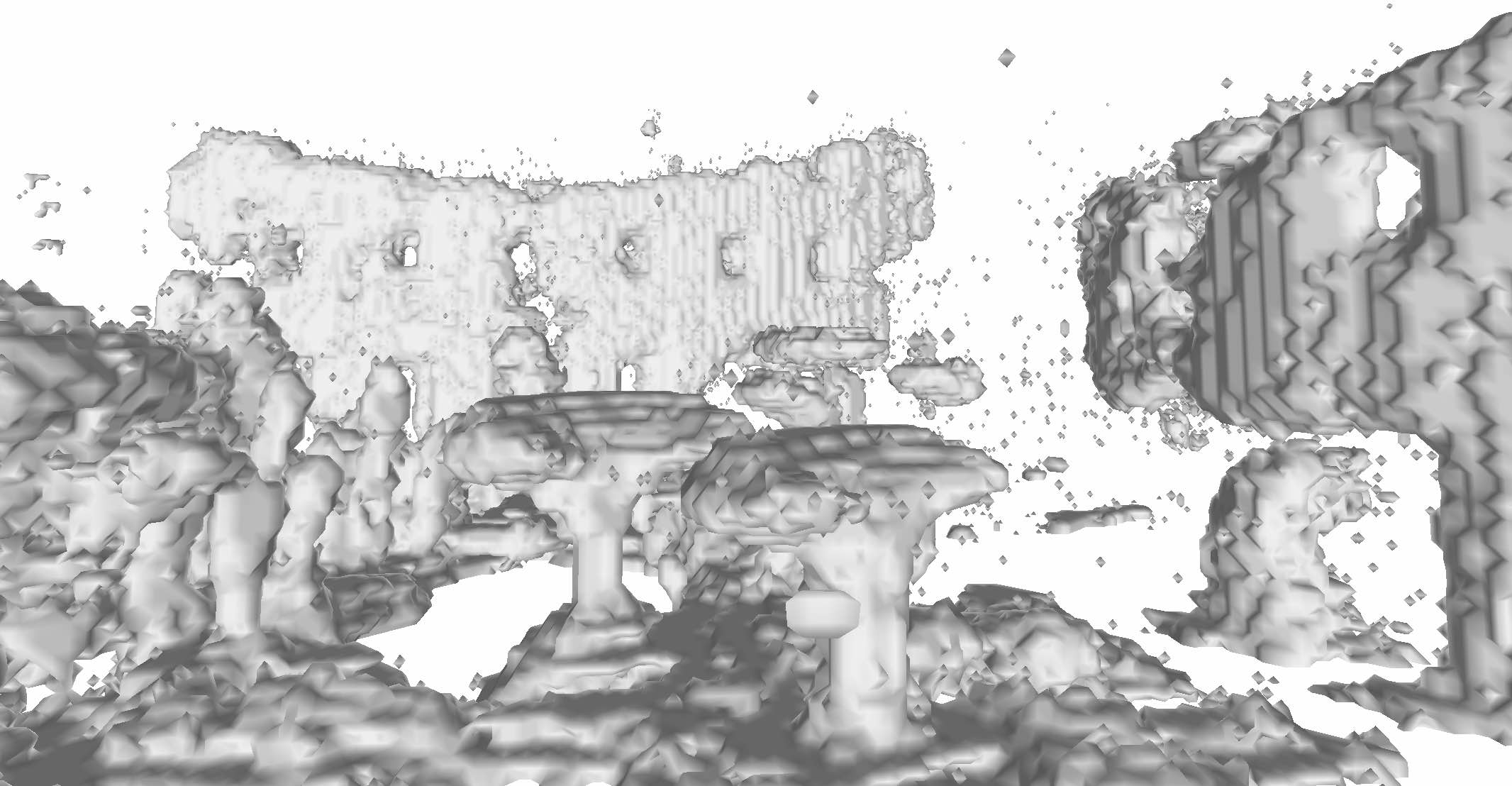} & 
\includegraphics[width=\ethWg\textwidth]{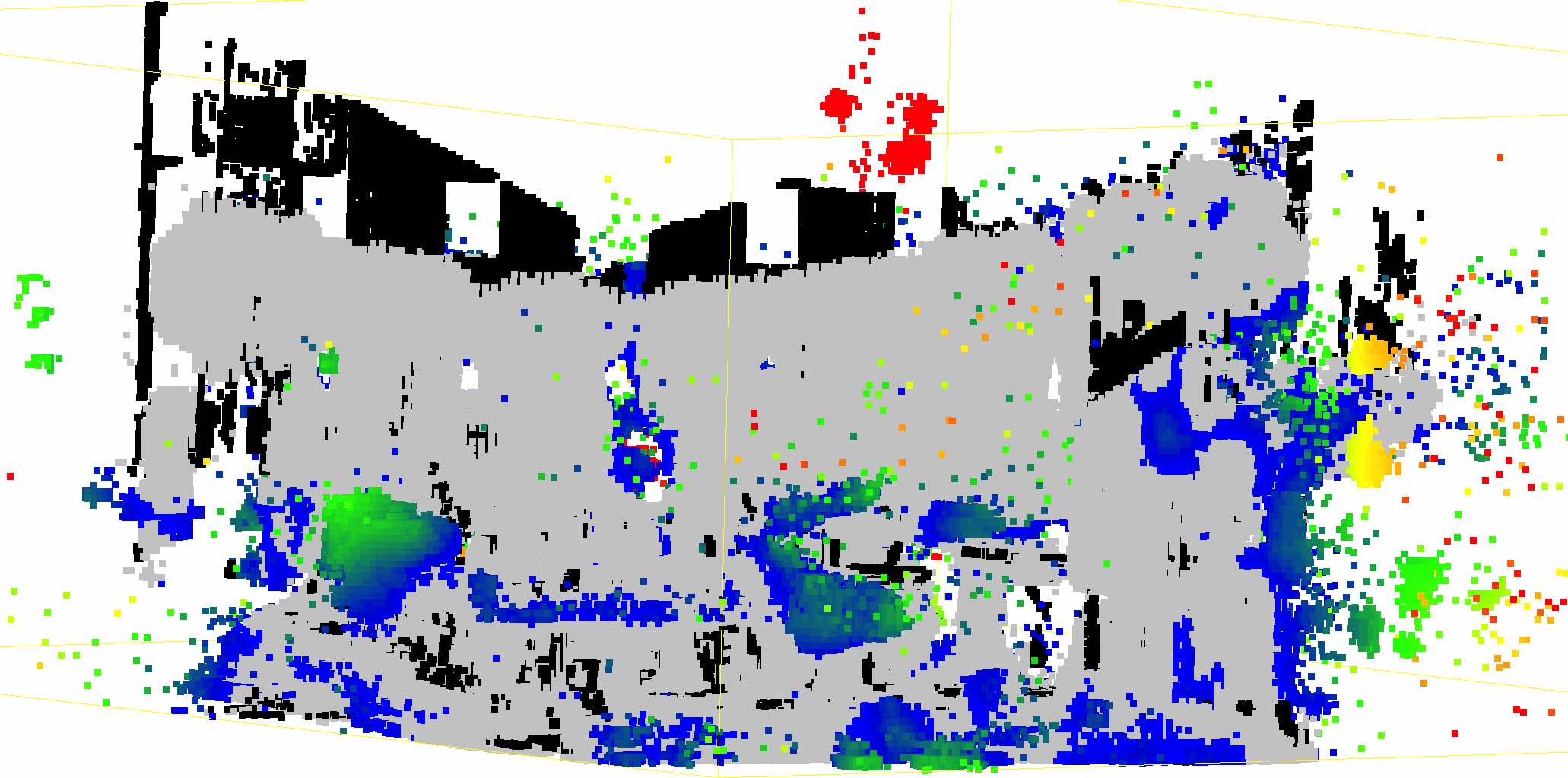} \\

Ground Truth & Standard~\cite{Cherabier-et-al-ECCV-2018} (Geometry) & Standard~\cite{Cherabier-et-al-ECCV-2018} (Error) & Learned (Geometry) & Learned (Error) \\[4pt]
\end{tabular}
\caption{\textbf{Experiment on ETH3D}. Two top rows contain the first training scene~--~\textit{delivery area}, and two middle rows another training scene~--~\textit{terrains}. Two bottom rows contain the validation scene~--~\textit{playground}. For each scene we show a full view and a close-up view. The distance to the ground truth (error) is color-coded by a gray color where the distance is less than 5 voxels (correct reconstruction) and blue-green-yellow-red color scale for outliers. 
Black voxels denote regions where the ground truth geometry was not reconstructed.}
\label{fig:eth3d}
\end{figure*}

%% file: figs/scannet_eth3d_table.tex
\begin{table}
    \centering
    \scriptsize
    \setlength{\tabcolsep}{11pt}
    \begin{tabular}{lcccc}
         \toprule
       Method & TP rate & Distance & SA & FA \\  \hline
       Input & 0.507 & 3.376 & 0.55 & 0.79 \\
       ScanComplete~\cite{Dai-et-al-CVPR-2018} & 0.588 & 2.527 & 0.47 & 0.90 \\
       Standard TSDF in~\cite{Cherabier-et-al-ECCV-2018} & 0.837 &  1.606  &  0.79 &  0.96   \\
       Proposed & {\bf 0.953} & {\bf 1.410}  & {\bf 0.90} & {\bf 0.97} \\
      \bottomrule
    \end{tabular}
    \vspace{3pt}
    \caption{\textbf{Quantitative results on the ScanNet dataset~\cite{scannet}.} We report true positive (TP) rate of completion, average surface distance to the ground truth, semantic accuracy (SA) and free-space accuracy (FA).
    The comparison demonstrates that our method performs better than the baselines~\cite{Cherabier-et-al-ECCV-2018,Dai-et-al-CVPR-2018}.
    }
    \label{tab:completion}
\end{table}

\begin{table} 
\centering
\setlength{\tabcolsep}{3.8mm}
\begin{tabular}{cccc}
\toprule
Dataset & Fusion Method &  SA & FA \\
\midrule
\multirow{2}{*}{ETH3D} & Standard TSDF in~\cite{Cherabier-et-al-ECCV-2018} &  0.50  &  0.96  \\
 & Learned (proposed) & \textbf{0.59} &  \textbf{0.97} \\
\bottomrule
\end{tabular}
\vspace{0.2em}
\caption{\textbf{Evaluation on the ETH3D dataset~\cite{eth3d}.} We report semantic accuracy (SA) and free-space accuracy (FA). The proposed learned fusion outperforms the baseline~\cite{Cherabier-et-al-ECCV-2018} with standard TSDF fusion, when used with different sensors.}\label{tab:scannet_eth3d}
\end{table}


%% file: sec/ex_scannet.tex
Previous two experiments were done on a synthetic dataset. The next evaluation on ScanNet~\cite{scannet} dataset shows that the method is able to perform well also on real data. 
However, the dataset contains only measurements from one sensor, Kinect. In order to create an additional sensor, we simulated an artificial noisy Kinect with outliers modeled by Gaussian noise with zero mean and standard deviation of 2 meters with probability of 1\%. We used 7 training scenes and 5 validation scenes from the hotel bedroom category, which have 9 semantic labels. Ground truth was obtained by running total variation on all views, whereas only every 10th view was used for further fusion. The proposed fusion method was compared to two state-of-the-art baselines, Cherabier~\etal~\cite{Cherabier-et-al-ECCV-2018} with simple averaging of TSDF volumes and ScanComplete~\cite{Dai-et-al-CVPR-2018}. ScanComplete also optimizes geometry, but is not designed for sensor fusion. Hence, this method performs worse when we input uniformly averaged multi-sensor data. ScanComplete is trained on SUNCG and fine-tuning on ScanNet is difficult due to incompleteness as already stated by the authors and thus omitted.

Tab.~\ref{tab:completion} shows various performance scores of the input geometry in comparison to the completed results. The proposed learned fusion improves semantic accuracy of ~\cite{Cherabier-et-al-ECCV-2018} by 11\%. Volumes with estimated confidence values are visualized in Fig.~\ref{fig:scannet}, together with two selected reconstructed validation scenes. Learned confidence values in the top row show that the network is able to learn different weights for artificially created noisy Kinect sensor and downscales occasional noisy pixels. Voxels outside walls are not downscaled and not penalized, because they are part of the \textit{unknown} label which is not included in the loss function. For the non-noisy Kinect, the confidence values are decreasing for voxels further away from the center. 
The Kinect sensor is known to have less precise measurements with increasing depth~\cite{Zeisl-Pollefeys-ICRA-2016} and this was learned by the network.

%% file: sec/ex_eth3d.tex
The last experiment on ETH3D dataset is done to confirm that the proposed method is able to work not only on real data, but also with several real sensors. ETH3D dataset~\cite{eth3d} comprises multi-view images with high resolution camera, as well as with low resolution camera rigs. The ground truth is given by a laser scan. 

We again tested that the joint learned fusion performs better than~\cite{Cherabier-et-al-ECCV-2018} with simple TSDF averaging. Training set had two scenes, \textit{delivery area} and \textit{terrains}, whereas validation set consisted of a single scene \textit{playground}. 
Only these three scenes contain measurements from both sensors, which explains the used set size. 
The resolution was set to 8 cm, which gives scenes large enough for training. The number of parameters to learn is low and the results show that only several scenes are enough to train the model. The label set consists of only two labels, \textit{free-space} and \textit{occupied-space}, as no semantic ground truth is available. 

Tab.~\ref{tab:scannet_eth3d} contains quantitative results on ETH3D dataset, where the increase in semantic accuracy is 9\%. Fig.~\ref{fig:eth3d} shows visualized reconstructions of all scenes with one close-up view for each scene. The learned fusion is able to provide more complete reconstructions and it does not contain as many separate outlier semantic voxels in ground truth free-space. The error is measured as distance to the ground truth with gray color regions representing the correct reconstruction with error less than 5 voxels.

%% file: sec/conclusion.tex
We proposed a novel machine learning-based depth fusion method that unifies semantic depth integration, multi-sensor or multi-algorithm data fusion as well as geometry denoising and completion.
We substantially generalize the recent semantic 3D reconstruction method~\cite{Cherabier-et-al-ECCV-2018} to incorporate an arbitrary amount of depth sensors.
To balance the contribution of each sensor according to their noise statistics, we extract features from the sensor data and learn the network to predict suitable confidence weights for each sensor and each point in space.
Our approach is generic and can also learn reliability statistics of different stereo algorithms. 
This allows us to use the method as an expert system that weights and fuses the outputs of all algorithms, providing a result that is better than of any individual algorithm.